\documentclass[acmtog]{acmart}

\acmSubmissionID{288}

\usepackage{booktabs} 

\usepackage{appendix}

\usepackage{animate}
\graphicspath{ {./figures/} }
\usepackage{cleveref}
\newcommand{\tabref}[1]{\tablename~\ref{#1}}
\newcommand{\figref}[1]{\figurename~\ref{#1}}
\usepackage{comment}
\usepackage[normalem]{ulem}
\DeclareGraphicsExtensions{.jpg, .pdf, .png, .gif}
\usepackage{soul}

\usepackage[ruled]{algorithm2e} 

\SetAlFnt{\small}
\SetAlCapFnt{\small}
\SetAlCapNameFnt{\small}
\SetAlCapHSkip{0pt}

\acmJournal{TOG}

\setcopyright{acmlicensed}\acmJournal{TOG}
\acmYear{2020}\acmVolume{39}\acmNumber{6}\acmArticle{225}\acmMonth{12} \acmDOI{10.1145/3414685.3417826}

\begin{document}
\title{Face Identity Disentanglement via Latent Space Mapping}

\author{Yotam Nitzan}
\affiliation{%
 \institution{Tel-Aviv University}
 }
\email{yotamnitzan@gmail.com}
\author{Amit Bermano}
\affiliation{%
 \institution{Tel-Aviv University}
}
\author{Yangyan Li}
\affiliation{%
    \institution{Alibaba Cloud Intelligence Business Group}
}
\author{Daniel Cohen-Or}
\affiliation{%
 \institution{Tel-Aviv University}
}

\begin{abstract}
\label{sec:abstract}

Learning disentangled representations of data is a fundamental problem in artificial intelligence.
Specifically, disentangled latent representations allow generative models to control and compose the disentangled factors in the synthesis process. 
Current methods, however, require extensive supervision and training, or instead, noticeably compromise quality. 

In this paper, we present a method that learns how to represent data in a disentangled way, with minimal supervision, manifested solely using available pre-trained networks. Our key insight is to decouple the processes of disentanglement and synthesis, by employing a leading pre-trained unconditional image generator, such as StyleGAN. By learning to map into its latent space, we leverage both its state-of-the-art quality, and its rich and expressive latent space, without the burden of training it.

We demonstrate our approach on the complex and high dimensional domain of human heads. We evaluate our method
qualitatively and quantitatively, and exhibit its success with de-identification operations and with temporal identity coherency in image sequences. Through extensive experimentation, we show that our method successfully disentangles identity from other facial attributes, surpassing existing methods, even though they require more training and supervision. 

\end{abstract}



\maketitle

\section{Introduction}
\label{sec:Intro}

Since the dawn of machine learning, learning a disentangled representation has been one of its  core problems. Disentanglement can be defined as the ability to control a single factor, or feature, without affecting other ones \cite{locatello2018challenging}. A properly disentangled representation can benefit semantic data mixing \cite{johnson2016perceptual, xiao2019identity}, transfer learning for downstream tasks \cite{bengio2013representation, tschannen2018recent}, or even interpretability \cite{mathieu2018disentangling}. Achieving disentanglement, however, is a notoriously difficult task, which has been addressed by many approaches. 

A key challenge to learning a disentangled representation is reducing supervision. Similarly to many other learning-related objectives, fully supervised solutions are most effective \cite{aberman2019learning, reed2015deep}, but impose often infeasible data collection requirements. It is not tractable, for example, to find a data set of paintings depicting the same scenes in different styles. The opposite approach, of a completely unsupervised disentanglement, is equally impractical at the moment, as it typically struggles with producing satisfying results \cite{locatello2018challenging}. Therefore, middle ground forms of supervision have been proposed. A prominent example is  
\textit{class-supervision}, where only the feature of interest is labeled throughout the dataset, partitioning it into classes. The class-supervised setting assumes the existence of multiple samples in each class, and that the intra-class variation of this feature is significantly lower than the inter-class ones  \cite{gabbay2019demystifying}. While being more feasible, this approach still requires meticulous gathering and labeling of data. Avoiding the labeling requirement would enable using virtually endless amounts of data.

\begin{figure}[t]
\centering
\includegraphics[width=\linewidth]{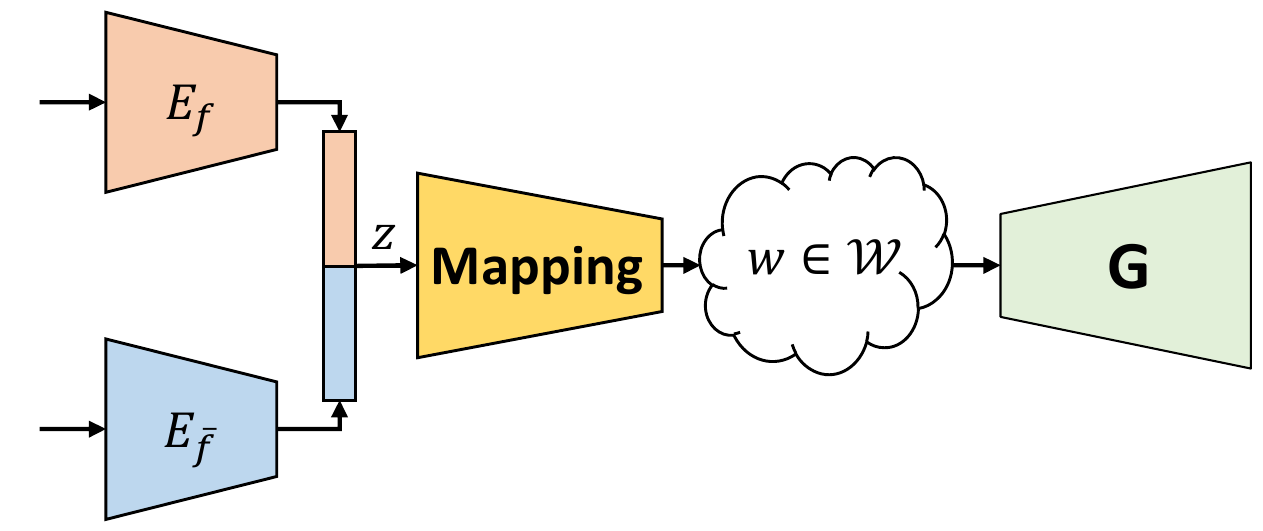}
\caption{Our disentanglement framework uses two encoders (left) to generate the latent code $z$, consisting of a description of the property of interest ($f$), and all the rest ($\overline{f}$). The code is then mapped to the latent space $\mathcal{W}$ of the employed pre-trained generator $G$. This decouples the tasks of learning quality image generation and disentanglement.}
\label{fig:teaser}
\end{figure}

\begin{figure}[t] 
\centering
\includegraphics[width=\linewidth]{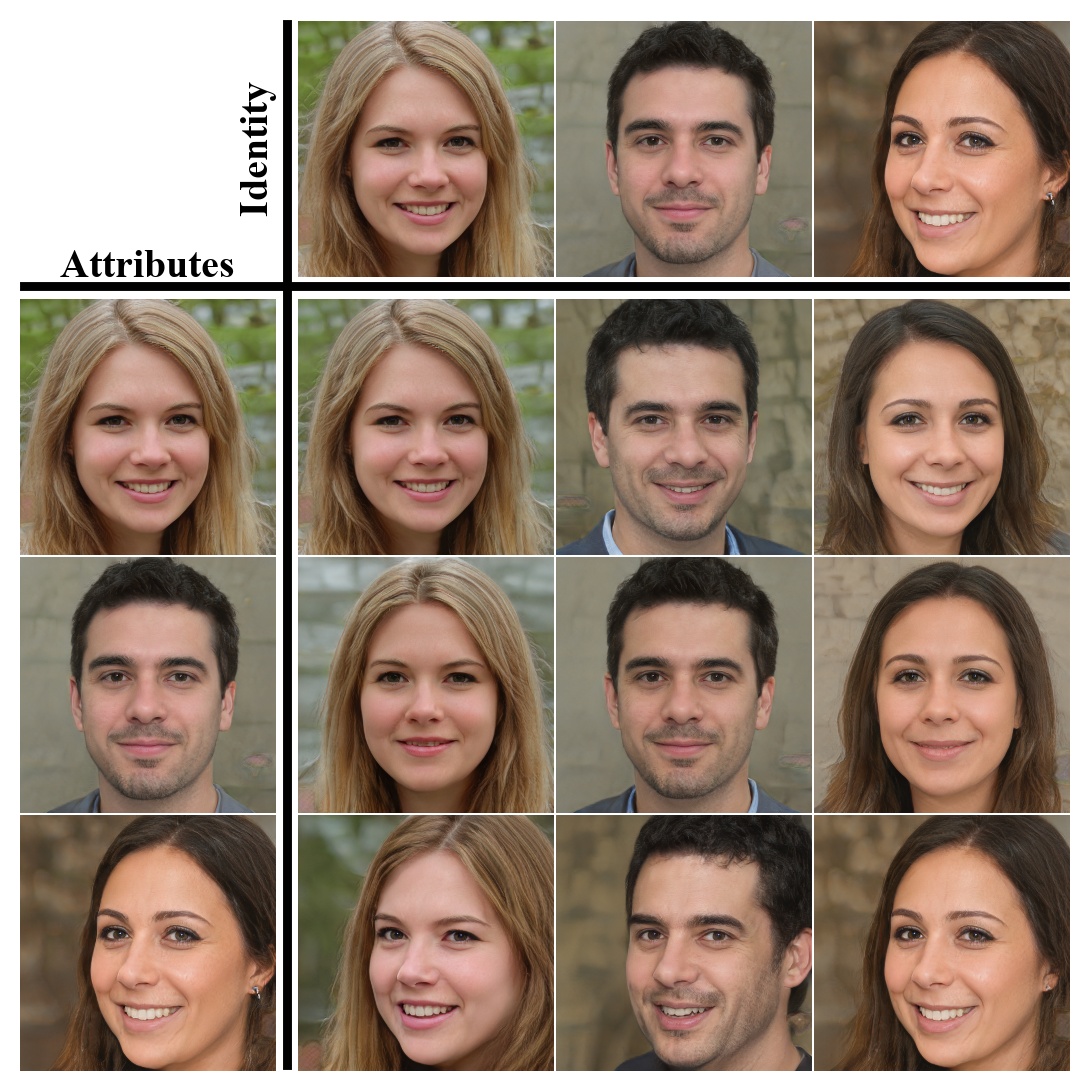}
\caption{Sample results generated by our method, demonstrating the ability to disentangle identity from other facial attributes: pose, expression and illumination and preserving one while manipulating the other. Three images are used as input, forming a 3 by 3 table combinations generated by our method. As can be seen, identity is preserved along the columns, and attributes are preserved along the rows. }
\label{fig:3x3_table_results}
\end{figure}

In this paper, we present a novel method to disentangle face identity from all other facial attributes, using no data specific supervision.
In our case, supervision is solely realized through the use of relevant pre-trained networks --- a plausible prerequisite, as shall soon be demonstrated, especially since these networks need not be trained on the same dataset or for the same task. Our key idea is to directly map the disentangled latent representation to the latent space of a pre-trained generator, as depicted in \figref{fig:teaser}. Given a feature of interest $f$, identity in our case, we propose training two encoders, as is commonly done in disentanglement setting \cite{bao2018towards, gabbay2019demystifying}, $E_f$ and $E_{\overline{f}}$, seeking to encode only $f$, and everything but $f$, respectively. Unlike traditional methods, however, we then propose to map the resulting latent code $z$ to the latent space $\mathcal{W}$ of a powerful, pre-trained generator $G$, and assess the quality of the disentanglement only on the latter's output. This mapping is the heart of our approach. It allows us to use a state-of-the-art pre-trained generator, inheriting its high-quality and fidelity, and to control its output in a disentangled manner with minimal training. Furthermore, our approach relaxes the requirement for a distinct disentanglement, where the representation is split into two parts which are mutually exclusive, and carry completely separate information. In our approach, the mapping is trained to disentangle and extract the relevant information from each part to be combined into a complete representation of the target image. In practice, this approach decouples the disentanglement task from the synthesis one, allowing the native employment of the most expressive and high-quality image generation techniques, and a dedicated training process for content control without compromising generation quality. 
We demonstrate our approach using arguably the most powerful unconditional image generator available nowadays --- StyleGAN \cite{karras2019style}, in one of the most challenging image generation domains --- the human face and head. Generating and manipulating faces is highly applicable on one hand, but is also known to be particularly hard, on the other. Besides the challenges of dealing with human faces that arise from the keen human perception of them \cite{mori1970uncanny}, and their high photometric, geometric and kinematic complexities, the human face has many independent, high dimensional attributes. From these, we choose to demonstrate image synthesis with disentangled control over the person identity attribute, as illustrated in \figref{fig:3x3_table_results}. This type of control is useful in applications such as de-identification, reenactment, and many others. Unlike other methods \cite{gabbay2019demystifying, bouchacourt2018multi, hadad2018two, denton2017unsupervised}, our training data does not contain examples of the same person twice, nor does it have any indication or labeling regarding the person's identity. Through the use of available networks for evaluating identity and facial landmarks, our approach effectively transforms StyleGAN into a conditional image generator, conditioned on either the identity of the person or all other facial attributes, such as expression, pose and illumination. 

As we shall see, the performance of our disentangled image generation heavily depends on the capabilities of the selected generator. In the case of StyleGAN, this means phenomenal image quality, outperforming all previous disentangled control attempts we compare to, but at the cost of expressiveness, as some of the faces do not reside within the attainable domain of the generator, due to the data used during its training. Nevertheless, 
in addition to superior quality, our method also successfully handles the generation of the entire head, including the hair --- a region that is known to have a strong impact on identity \cite{abudarham2019critical}. This is in contrast to state-of-the-art identity manipulation methods, which manipulate facial features only  ~\cite{bao2018towards, li2019faceshifter, nirkin2019fsgan,gafni2019live}.

As validation, we offer several experiments, evaluating qualitatively and quantitatively face identity and attributes manipulations, and compare them to previous methods. The experiments assess feature combination, de-identification operations, and temporal coherency of identity over sequences. The methods we compare to include state-of-the-art class-supervised disentanglement methods, which rely on a more structured, better labeled, but harder to curate, data. 
We evaluate all methods in terms of the quality of disentanglement and preservation of composing factors on unseen faces, as well as image quality and diversity. Our method is shown to outperform previous art, in addition to offering unique advantages, such as the said generation of the entire head and hair, and minimal supervision, which does not necessitate multiple images of the same person at any point.

\section{Related work}
\label{sec:related}

\subsection{Disentanglement}
Many works learn disentangled representations and use various levels of supervision. Fully supervised methods \cite{aberman2019learning, reed2015deep} learn from a dataset in which all relevant underlying factors are labeled. In this case, a sample from the dataset takes the form of \textit{(input, transformation, result of transformation)}. Critically, a ground truth is available for any transformation performed on the data. This data requirement is infeasible in many domains and tasks. On the other end of the spectrum, fully unsupervised methods \cite{higgins2017beta, kim2018disentangling, chen2016infogan} learn from a dataset with no associated information. These methods employ information-theoretic regularization losses to encourage disentanglement. These methods trade-off quality for disentanglement, often explicitly \cite{higgins2017beta, kim2018disentangling}, and thus produce low visual quality results.
Class-supervised methods \cite{bao2018towards, gabbay2019demystifying, bouchacourt2018multi, hadad2018two, denton2017unsupervised, xiao2019identity, li2019faceshifter} use additional labels that partition the images to classes, defined by a set of mutually exclusive attribute values. For example, each class contains a set of images of a single identity only, while other attributes vary. 
In contrast, we examine a more challenging but easily attainable dataset, in which no person appears twice, and no labeling is offered.

All aforementioned fully-supervised and class-supervised methods follow a similar approach, where separate encoders infer latent representations of their inputs. Next, those representations are fed into a generator that produces the output image. As described in Section \ref{sec:Method}, our choice of architecture follows the same approach. Among these works, perhaps the most similar to ours are the ones presented by Bao et al. \shortcite{bao2018towards} and its successor, FaceShifter \cite{li2019faceshifter}. Both focus specifically on disentangling identity from other attributes and use a pre-trained face recognition network as an encoder to infer the identity representation. There are several notable differences between these methods and ours.
Both methods train a generator themselves, while we use a pre-trained generator. Additionally, both methods deal with inner-facial features only, not including the head and hair.
Most importantly, both are class-supervised, explicitly requiring identity labels.
We qualitatively compare our method with FaceShifter in Section \ref{subsec:compare_previous}.

\subsection{Latent Space of GANs}
With the rapid evolution of GANs, many works have tried to understand and control their latent space.
Several methods apply \textit{GAN Inversion}, where the latent vector that most accurately reconstructs a given image is sought. Some methods train an encoder to map images to the latent space in conjunction to training the generator (e.g., \cite{pidhorskyi2020adversarial, huang2018introvae}). Training the generator together with the encoder requires long training and often compromises the generated images' quality. Therefore, other methods consider inversion as a post-hoc task where the generator is pre-trained and then inversion is solved separately. Such methods either directly optimize the latent vector to minimize reconstruction error for every image \cite{creswell2018inverting, lipton2017precise, abdal2019image2stylegan}, train an encoder to map images to the latent space \cite{luo2017learning, perarnau2016invertible, richardson2020encoding}, or use a hybrid approach combining both \cite{zhu2016generative, zhu2020domain, pbaylies-stylegan-encoder}.
A separate line of research deals with learning to traverse the latent space in a semantically meaningful manner. A popular approach is to find linear directions that correspond to changes in a given binary labeled attribute, such as young $\leftrightarrow$ old, or no-smile $\leftrightarrow$ smile \cite{denton2019detecting, shen2019interpreting, goetschalckx2019ganalyze}.
Tewari et al. \shortcite{tewari2020stylerig} utilize a pre-trained 3DMM to learn semantic face edits in latent space.
Jahanian et al. \shortcite{jahanian2019steerability} find latent space paths that correspond to a specific image transformation, such as zoom or rotation, in a self-supervised manner. Härkönen et al. \shortcite{harkonen2020ganspace} find useful paths in a completely unsupervised manner. These paths are set to be the principal component axes (PCA) on an intermediate activation space.   
Similar to the other methods, the computed transitions control one-dimensional attributes such as age or gender, as well as image transformations like zoom or rotation.

Our work is different from the two lines of work discussed above. First, inversion methods start from a single image and aim to infer a latent vector that reconstructs it. This means that the input image serves as a particularly strong, pixel-level, supervision for the expected output. In contrast, our method infers a latent vector that represents an unseen image, and therefore cannot rely on pixel-level supervision, but on a significantly weaker form of supervision.
The second difference concerns image manipulation. We perform manipulation by directly mapping properties from two given images to a latent vector that represents the unseen output comprising both properties. This is in contrast to previous methods that propose traversing the latent space to perform image editing. 

It may be possible to devise a way to apply inversion followed by a traversal of the latent space to perform a task similar to the one we describe, i.e., to produce a novel image that keeps properties from one image, and alters some other ones. However, all properties we generate seek to match those of the two input images, i.e., identity from one image and facial attributes from the other. On the other hand, state-of-the-art methods based on image inversion  cannot achieve the same goal. For example, they cannot generate an image featuring the identity of the input image, but also the smile of another specific one. Such an approach would require traversing the latent space, which renders mimicking a specific expression extremely challenging.
Furthermore, latent traversal methods depend more on the local behavior of the latent space, assuming it is well-behaved everywhere. Our method, on the other hand, bypasses the need to explicitly invert an image but rather produces the output image through a direct mapping.

\subsection{Controlling Facial Attributes}
There is an abundance of works on face manipulation with various means of controlling the facial attributes. These can be categorized by whether they preserve identity and manipulate other attributes or vice versa.

As for the former, some works are based on conditional GANs that translates between different domains such as young $\rightarrow$ old, or happy $\rightarrow$ angry ~\cite{choi2018stargan, choi2019stargan, perarnau2016invertible, liu2017unsupervised, pumarola2018ganimation}. These methods are limited to a discrete number of domains and require datasets with their associated labels, but work on unseen image. By comparison, face reenactment methods transfer expression and head pose (sometimes also illumination and eye gaze) from a source video to a target identity ~\cite{kim2018deep, thies2016face2face, thies2018headon, elor2017bringingPortraits, zakharov2019few}. But, there methods usually require training a network for each given identity, and assumes the availability of videos of it.
Another line of identity-preserving works ~\cite{shen2018faceid, liu2018exploring, tian2018cr} train GAN-based Image-to-Image translation networks that preserve the identity of the input image, while a subset of other attributes are controlled by a different image. Unlike our method, these methods are all class-based, i.e. relying on having an identity-labeled image dataset with multiple images for each identity.

Face de-identification methods are of the latter category, editing the identity of a target image, while preserving other attributes, like expression, pose, illumination, etc. Sun et al. \shortcite{sun2018hybrid} remove the face from the target image and complete it using a GAN , while other methods ~\cite{gafni2019live, wu2018privacy} shift the identity using a pre-trained face recognition network, while keeping the general image relatively similar to the source.
A popular approach is performing face swap ~\cite{nirkin2019fsgan, li2019faceshifter, deepfake, bao2018towards}, in which the face in the target image is replaced by the one from the source image. Unlike previously mentioned methods, face swap allows controlling the generated identity which is copied from the source image. However, face swapping methods are generally limited to using a target face relatively similar to the source face.

Our method, allows the editing of both identity and facial attributes, while also controlling the generated identity according to an input image. In contrast to other methods, our method is successful even when the input images are completely different, demonstrating our disentanglement capability.
Furthermore, all aforementioned methods only alter the internal features of the face, usually by cropping a tight face region or by using a segmentation mask of the face, leaving the head and hair intact. This approach conflicts with the fact that numerous works have identified that the appearance of the head as a whole, and specifically the hair, are crucial for identification ~\cite{sinha1996think, toseeb2012significance, abudarham2019critical, sendik2019s}. Our method, on the other hand, generates an entire human head, including the hair, thus better controlling and preserving the generated identity.

\newcommand{\norm}[1]{\left\lVert#1\right\rVert}

\section{Method}
\label{sec:Method}

\begin{figure*}[t]
	\centering
	\includegraphics[width=\linewidth]{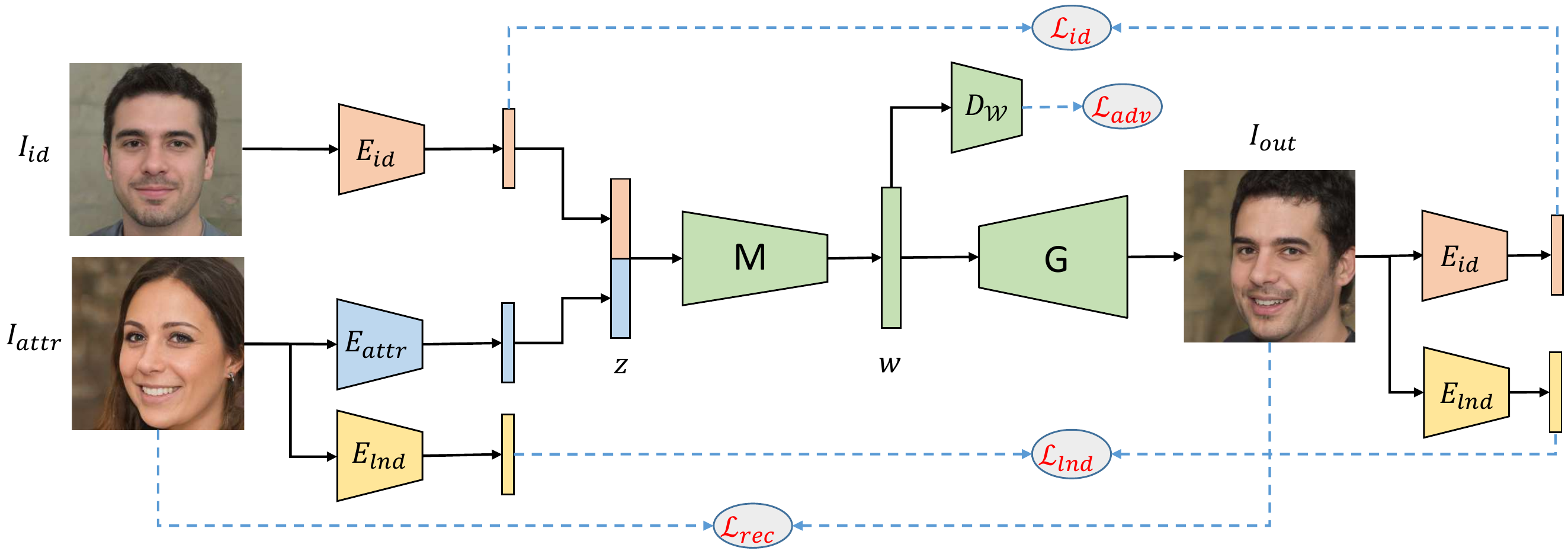}
	\caption{Our disentanglement scheme, as utilized in the human head domain. Data flow is marked by solid lines, and losses by dashed ones. The identity and attributes codes are first extracted from two input images using encoders $E_{id}$ and $E_{attr}$ respectively. Through our mapping network $M$, the concatenated codes are mapped to $\mathcal{W}$, the latent space of the pre-trained generator $G$, which in turn generates the resulting image.
	An adversarial loss $\mathcal{L}_{adv}$ ensures proper mapping to the $\mathcal{W}$ space. Identity preservation is encouraged using $\mathcal{L}_{id}$, that penalizes differences in identity between $I_{id}, I_{out}$. Attributes preservation is encouraged using $\mathcal{L}_{rec}$, $\mathcal{L}_{lnd}$, that penalizes pixel-level and facial landmarks differences respectively, between $I_{attr}, I_{out}$.}
	\label{fig:arch}
\end{figure*}

Our method takes two images as input: $I_{id}, I_{attr}$. The goal is to generate an image with the identity from $I_{id}$ and all other attributes, specifically pose, expression and illumination from $I_{attr}$. The disentanglement task is therefore to disentangle identity from all other attributes. Once achieved, We can extract the identity from $I_{id}$ and the attributes from $I_{attr}$, and reassemble them into a combined representation of a new human head. This representation is then fed to the generator, to generate an image that respects both the identity of $I_{id}$ and attributes of $I_{attr}$. 
Encouraging this disentanglement, while generating state-of-the-art quality images, is notoriously difficult. The key idea of our work is to separate the two objectives, by learning to map the combined representation into the latent space of a pre-trained generator, and its existing semantics.
For this reason, we use a pre-trained generator with semantics-rich, and expressive latent space.

As depicted in \figref{fig:arch},
the network consists of two encoders $E_{id}$ and $E_{attr}$, a mapping network $M$, and a generator network $G$. An additional encoder, $E_{lnd}$ and discriminator $D_\mathcal{W}$ are used for loss calculation only and shall be discussed in Section \ref{subsec:training_and_losses}.
The task of generating an image with the same identity as in $I_{id}$, and the attributes portrayed in $I_{attr}$, consists of two parts: extracting identity and attributes from the corresponding input images, and then reassembling them to create a new head representation and generating an image accordingly. For the former, we use the two encoders $E_{id}, E_{attr}$. For the latter, we combine the codes by concatenation:
\begin{equation}
    z = [E_{id}(I_{id}), E_{attr}(I_{attr})]
\end{equation}
and map $z$, using $M$, into the latent space of a pre-trained state-of-the-art generator $G$, which then generates the output image, $I_{out}$.
We use state-of-the-art StyleGAN \cite{karras2019style} as the pre-trained generator for all our experiments. Differently from other GANs, StyleGAN has two latent spaces: $\mathcal{Z}$, which is induced by a fixed distribution, and $\mathcal{W}$ induced by a learned mapping from $\mathcal{Z}$. We choose to map the combined face code into $\mathcal{W}$, as it is a more disentangled latent space than $\mathcal{Z}$, thus more suitable to facilitate and accommodate image editing \cite{karras2019style, shen2019interpreting}.
The design choice of using an existing latent space is crucial for a few reasons. Usually disentanglement is performed with the premise that given enough samples with constant factors, while other factors are varying, one could learn to identify the constant factors and disentangle it from the others e.g., LORD \cite{gabbay2019demystifying}.
In our setting, the dataset does not contain more than one image of the same person. Therefore, it is unclear how the network could learn to disentangle. Our approach resolves this problem by leveraging a latent space that already exhibits some degree of disentanglement, achieved in a completely unsupervised manner. Moreover, by using a state-of-the-art generator, we alleviate the difficulty of learning to generate high-quality and high-fidelity images. However, training the mapping between the latent space of the encoder and $\mathcal{W}$, is not trivial. Thus, we add a discriminator $D_{\mathcal{W}}$ to help $M$ predict features that lie within $\mathcal{W}$. $D_{\mathcal{W}}$ is trained in an adversarial manner to discriminate between real samples from StyleGAN's $\mathcal{W}$ space and $M$'s predictions.
Note that, thanks to our use of a pre-trained generator, there is no discriminator employed on $I_{out}$. Thus, side-stepping much of the difficulty of training adversarial methods.

\subsection{Network Architecture}
The $E_{id}$ encoder is a pre-trained ResNet-50 \cite{he2016deep} face recognition model, trained on VGGFace2 \cite{cao2018vggface2}, with loose crops including the hair. The $E_{attr}$ encoder is implemented as InceptionV3 \cite{szegedy2016rethinking}. For both encoders, their output is taken from the last feature vector before the FC classifier. The mapping network, $M$, is a 4-layers MLP with LReLU \cite{he2015delving} activation layers. The generator, $G$ is a pre-trained StyleGAN synthesis network, trained on FFHQ \cite{karras2019style}. $G$ takes our predicted $w$ vector as input and employs it normally through the AdaIN \cite{huang2017arbitrary} layers.
Both $E_{id}$ and $G$ are kept frozen during training, while all other networks are trainable.

\subsection{Training and Losses}
\label{subsec:training_and_losses}

We create a dataset using StyleGAN in the following manner. We sample 70,000 random Gaussian vectors and forward them through a pre-trained StyleGAN. In the forward process, the Gaussian noise is mapped into a latent vector $w$, from which an image is generated, and we record both the image and the $w$ vector.
The StyleGAN generated images are used as our training dataset, and the latent $w$ vectors are used as "real" samples for training $D_{\mathcal{W}}$. 

StyleGAN cannot create the entire human head space, specifically all human identities, from its latent space $\mathcal{W}$. Some works \cite{zhu2020domain, abdal2019image2stylegan} used an artificially enlarged latent space, named $\mathcal{W}+$, from which the generator may also create non-human images, including cats and bedrooms. We therefore choose to use a StyleGAN generated dataset to prevent the conflict between identity preserving and mapping into StyleGAN's rich latent space $\mathcal{W}$.

For adversarial loss, we use the non-saturating loss \cite{goodfellow2014generative} with $R_1$ regularization \cite{mescheder2018training}:
\begin{equation}
\begin{gathered}
\mathcal{L}_{adv}^{D} =  -\underset{w \sim \mathcal{W}}{\mathbb{E}}[log D_{\mathcal{W}}(w)]-\underset{z}{\mathbb{E}}[log (1-D_{\mathcal{W}}(M(z)))]  + \\ 
\frac{\gamma}{2} \underset{w \sim  \mathcal{W}}{\mathbb{E}} \left[ \norm{\nabla_w D_{\mathcal{W}}(w)}_2^2 \right]
\end{gathered}
\end{equation}

\begin{equation}
    \mathcal{L}_{adv}^{G} = - \underset{z}{\mathbb{E}}[log D_{\mathcal{W}}(M(z))]    
\end{equation}

An $L_1$ cycle consistency loss between $I_{id}$ and $I_{out}$ is used to enforce identity preservation:
\begin{equation}
\mathcal{L}_{id} = \norm{E_{id}(I_{id}) - E_{id}(I_{out})}_{1}
\end{equation}
As discussed, human perception is highly sensitive to minor artifacts in facial appearance, this is especially true when generating sequences of frames, where not only does every individual frame must look realistic, but the motion across frames must also be realistic. Facial landmarks model the possible motion of the human face, Therefore, we incorporate a sparse $L_2$ cycle consistency landmarks loss. Landmarks are extracted using a pre-trained network noted as $E_{lnd}$:
\begin{equation}
    \mathcal{L}_{lnd} = \norm{E_{lnd}(I_{attr}) - E_{lnd}(I_{out})}_{2}    
\end{equation}

Additional loss is exercised to encourage pixel-level reconstruction of $I_{attr}$.
This loss is clearly motivated by our desire for $I_{out}$ to be generally similar to $I_{attr}$. Intuitively, if $I_{id}, I_{attr}$ are the same image, we would expect our method to reconstruct this image. Furthermore, we would like to capture and preserve pixel-level information such as colors and illumination, not modeled by any other loss.
For this end, we adopt the "mix" loss suggested in Zhao et al. \shortcite{zhao2016loss}, and use a weighted sum of $L_1$ loss and MS-SSIM loss:
\begin{equation}
    \begin{gathered}
    \mathcal{L}_{mix} = \alpha (1 - \textnormal{MS-SSIM}(I_{attr}, I_{out})) +  (1 - \alpha) \norm{I_{attr} - I_{out}}_1     
    \end{gathered}
\end{equation}

However, a pixel reconstruction loss might also affect the identity of $I_{out}$ by reconstructing facial features from $I_{attr}$. In order to prevent this, we employ the reconstruction loss only when $I_{id} = I_{attr}$, i.e. :
\begin{equation}
\mathcal{L}_{rec} = \begin{cases}
\mathcal{L}_{mix}, & I_{id} = I_{attr} \\
0, & \text{Otherwise}
\end{cases}
\raisetag{2\normalbaselineskip}
\end{equation}

The overall generator non-adversarial loss is a weighted sum of the above losses:
\begin{equation}
\mathcal{L}_{non-adv}^{G} =\lambda_1 \mathcal{L}_{id} + \lambda_2 \mathcal{L}_{lnd} + \lambda_3\mathcal{L}_{rec}    
\end{equation}

Our training procedure is simple. We uniformly randomly sample images and latent vectors from our generated dataset. The images are used for $I_{id}, I_{attr}$ and the latent vectors are used as "real" samples for $D_{\mathcal{W}}$.
When $I_{id} \neq I_{attr}$, the network learns to disentangle identity from attributes. Whereas when $I_{id} = I_{attr}$ it learns to encode all the information needed for proper reconstruction.

\subsection{Implementation Details}
We use StyleGAN pre-trained at 256x256 resolution in all our experiments, to easily compare with other methods. Please see the supplementary material for 1024x1024 results as well. The ratio of  training samples with $I_{id} \neq I_{attr}$ and  $I_{id} = I_{attr}$ is an hyper-parameter that controls the weight for disentanglement and reconstruction.
We empirically take $I_{id} \neq I_{attr}$ every third iteration, and $I_{id} = I_{attr}$ otherwise.
$E_{lnd}$ is implemented using a pre-trained landmarks regression network \cite{feng2018wing}, trained to regress 68 facial keypoints.

We optimize the adversarial loss $\mathcal{L}_{adv}^{G}$ and non-adversarial losses $\mathcal{L}_{non-adv}^{G}$ separately, which proved to be more stable during training.
Training is performed using the Adam \cite{DBLP:journals/corr/KingmaB14} optimizer, with $\beta_1 = 0.9, \beta_2 = 0.999$.
We use learning rate of $5e^{-5}$ when optimizing $\mathcal{L}_{non-adv}^{G}$. For adversarial learning rates we follow Heusel et al. \shortcite{heusel2017gans} and use $5e^{-6}$ for $G$'s adversarial loss $\mathcal{L}_{adv}^{G}$ and $2e^{-5}$ for $D_{\mathcal{W}}$.
Loss weights are set to $\lambda_1 = 1, \lambda_2=1, \lambda_3 = 0.001, \lambda_4 = 0.02, \alpha=0.84$ and $\gamma = 10$.
When calculating $\mathcal{L}_{lnd}$ we use only the 52 inner-face landmarks, removing the jawline landmarks which were found to strongly affect the head shape, harming identity preservation.
The network is trained end-to-end with batch size 6 on a single NVIDIA Titan XP GPU and requires roughly a day to converge. Note that this is incredibly efficient, as training StyleGAN itself would require more than 30 days on the same GPU.
\section{Experiments}
\label{sec:experiments}
We perform extensive experimentation to evaluate our method, mainly through two aspects: the quality of the disentanglement, or how well we control identity and facial attributes without them affecting each other, and the quality of the synthesized images. We further perform ablation studies, which show the importance of individual components in our pipeline, and compare our performance to state-of-the-art methods.

\begin{figure*}[ht!]
	\centering
	\includegraphics[width=0.96\linewidth]{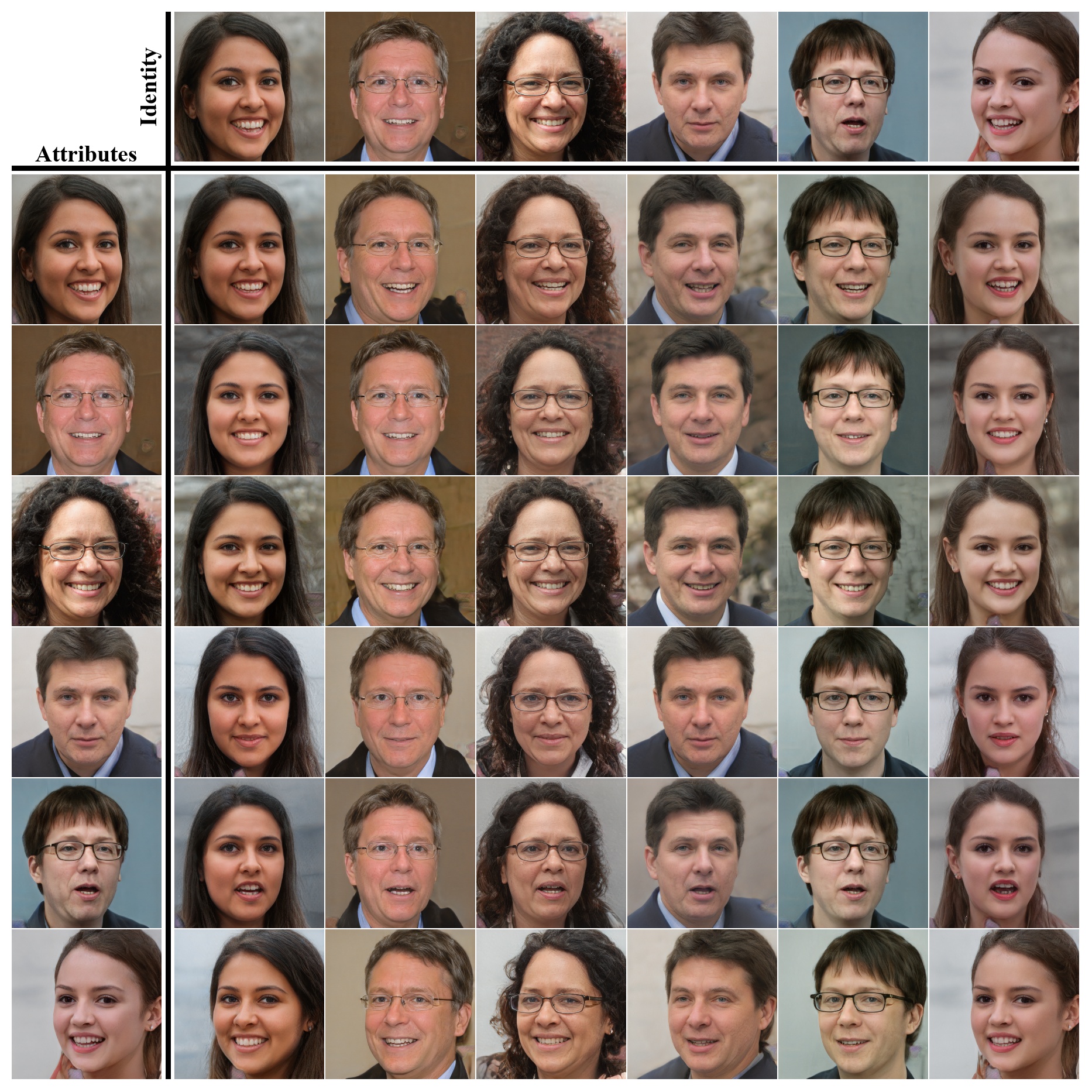}
	\caption{Feature combination results. For every image in the table, identity is taken from the top, and the rest of the attributes (including expressions, orientation, lighting conditions, etc.) from the left. All images (both inputs and output) were generated using StyleGAN. }
	\label{fig:table_results_1}
\end{figure*}

\begin{figure*}[ht!]
	\centering
	\includegraphics[width=0.96\linewidth]{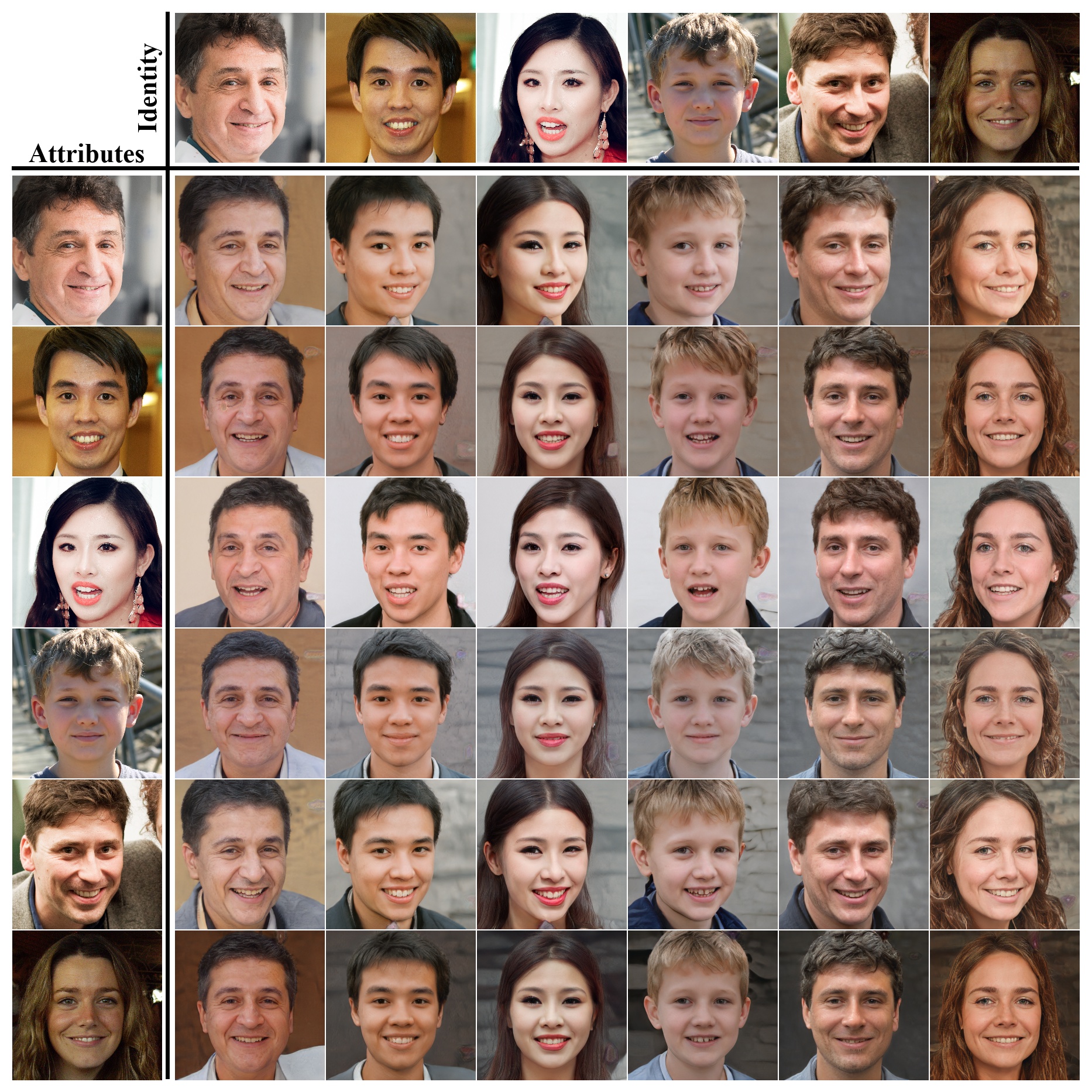}
	\caption{Feature combination results on FFHQ images. The setting is identical to \figref{fig:table_results_1}.}
	\label{fig:table_results_ffhq_1}
\end{figure*}

First, a qualitative inspection of the results are shown in \Cref{fig:table_results_1,fig:table_results_ffhq_1} for StyleGAN and FFHQ input images, respectively. The array of images illustrates the degree of identity preservation of our method (along the columns) and preservation of the rest of the attributes (along the rows). In addition, our method successfully preserves the overall head shape and the hair -- a pivotal yet elusive part of true identity preservation. Additionally, we observe consistency in details such as the existence and appearance of glasses. This is a crucial element when considering consistency, and is especially relevant when, for example, generating consecutive frames for a sequence. Note that these disentanglement and preservation capabilities cannot be achieved by the style mixing approach proposed in the original StyleGAN paper, where styles entangle identity and other semantic attributes.
A performance gap can be observed between our results on "real" images (FFHQ) and on StyleGAN generated images.
As previously discussed, we inherit the performance of the employed pre-trained generator and its latent space. As discussed in the literature \cite{abdal2019image2stylegan, zhu2020domain, karras2019analyzing}, StyleGAN is unable to generate the entire space of human heads from $\mathcal{W}$. 
This is most evident for the head pose, where faces generated by StyleGAN feature no roll angle, since the faces in the training data were vertically aligned. Similarly, not all human identities can be generated by StyleGAN. By inheriting StyleGAN's performance, our method generates the closest possible identity, which qualitatively and quantitatively is very similar as evident in \tabref{tab:Quantitative}.

\label{sec:table_results_1}

\label{sec:table_results_ffhq_1}

\subsection{Comparison with Previous Methods}
\label{subsec:compare_previous}

\begin{figure}[htp!]
	\centering
	\includegraphics[width=\linewidth]{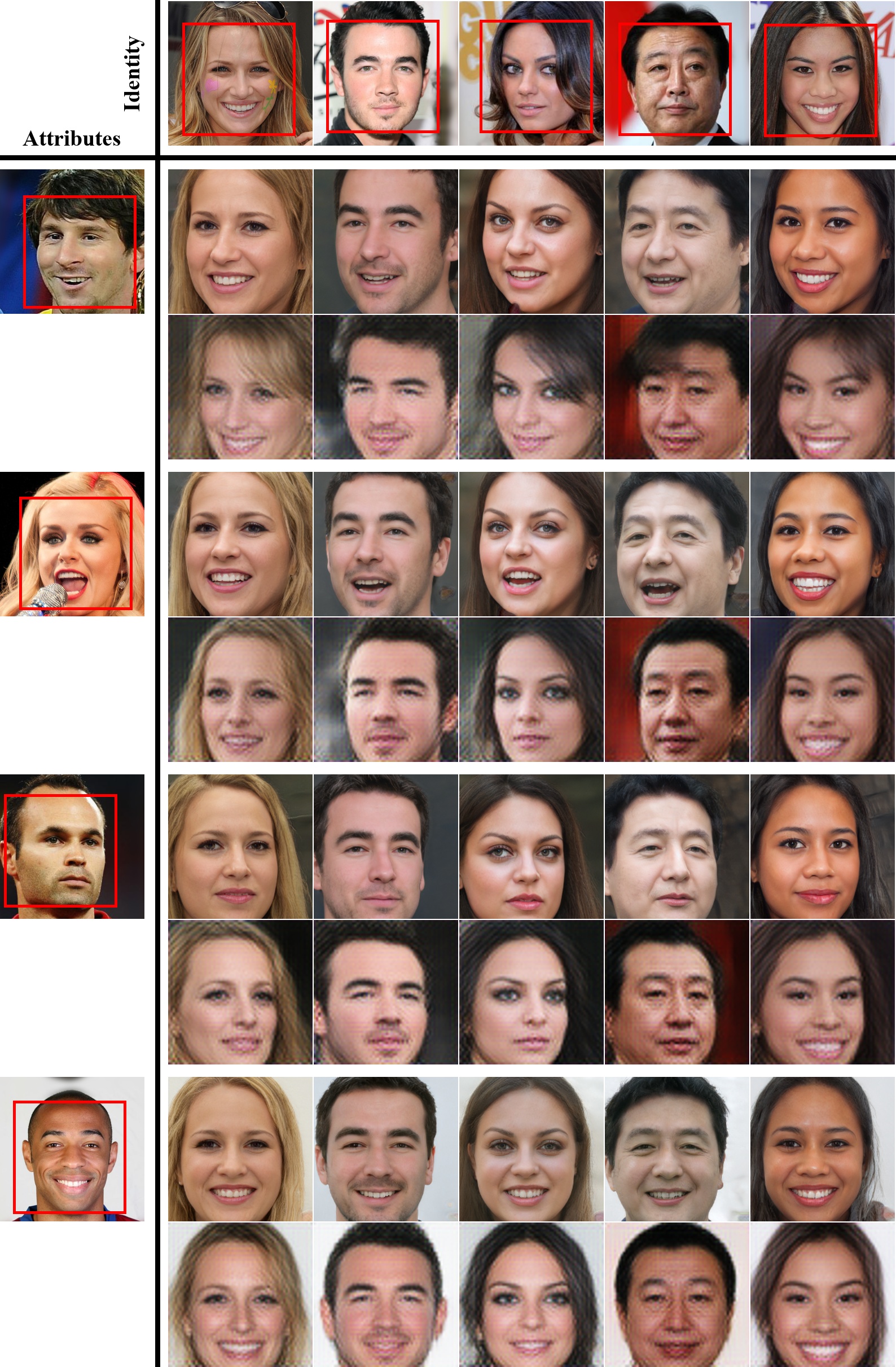}
	\caption{Qualitative Comparison of our method (odd rows) to LORD \cite{gabbay2019demystifying} (even rows) on samples from CelebA. Our results have a much better visual quality and preservation of identity and facial attributes (see \tabref{tab:Quantitative}). The red frame represents the region LORD gets as input.
	The reader is recommended to zoom in to better observe details.
	}
	\label{fig:compare_to_lord}
\end{figure}

We qualitatively compare our results against those of LORD \cite{gabbay2019demystifying} on images from CelebA \cite{liu2015faceattributes} in \figref{fig:compare_to_lord}. 
Note that the comparison is performed on images from the dataset on which LORD was trained. On the other hand, our method was trained on a significantly different dataset composed from StyleGAN generated images. Therefore, making this comparison significantly more challenging for our method.
LORD uses low resolution, 64x64 tight face crops, while our method handles higher resolution, of 256x256 loose crops. For a fair comparison, each method is run using its native input configuration, and in post-process we crop and resize the output images to make them visually comparable. The red frames indicate the region of the image that is input to LORD.

\label{sec:compare_to_lord}

\textfloatsep 5pt

As can be observed, our method better preserves the facial expression of the attributes image, regardless of the expression of the identity image, indicating a strong disentanglement between identity and expression. This is most noticeable when observing the mouth, where our method is able to generate various mouth shapes. On the other hand, LORD struggles in preserving expression when the identity image has a non-neutral expression, and preserving the challenging open-mouth expression. Furthermore, our results have a much higher visual quality than LORD's, which are of low resolution and contain artifacts, most noticeable is the checkerboard effect, which should not be attributed to the resizing of the image as it also exists in the original resolution.

\begin{figure}
	\centering
	\includegraphics[width=\linewidth]{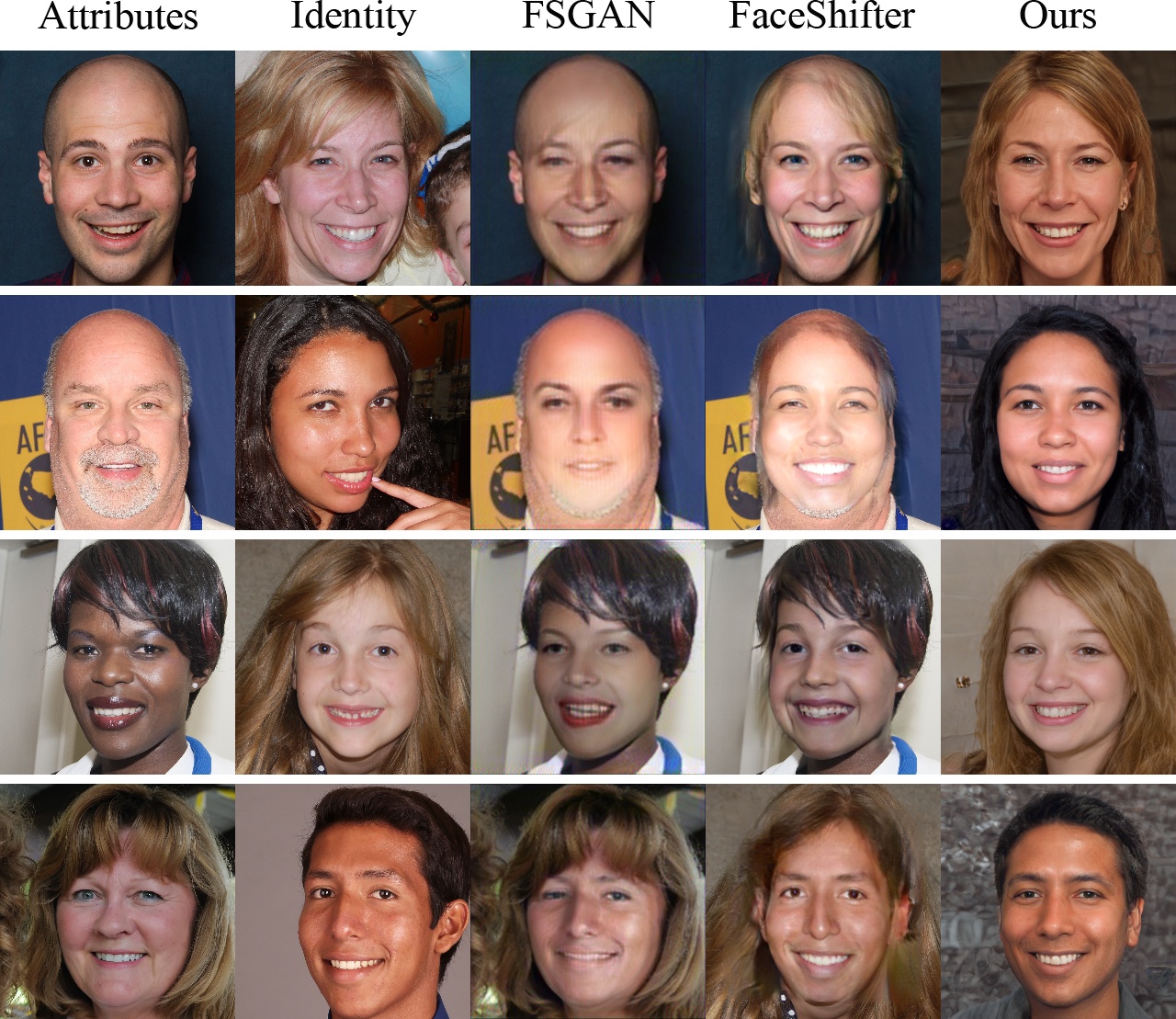}
	\caption{Qualitative comparison to state-of-the-art face swapping methods FaceShifter \cite{li2019faceshifter} and FSGAN \cite{nirkin2019fsgan} on samples from FFHQ \cite{karras2019style}. As can be seen, our method better preserves the identity as it does not only preserve inner facial features, but the entire head and hair, which are known to be crucial for identity recognition by humans. It can also be observed that face swapping methods struggle with faces with significantly different appearances.}
	\label{fig:compare_to_fs}
\end{figure}

We also compare our results with the latest face swapping methods FaceShifter \cite{li2019faceshifter} and FSGAN \cite{nirkin2019fsgan}. As discussed earlier, face swapping is a different yet related task that focuses on replacing inner face features only. We qualitatively demonstrate the differences in our application from face swapping in \figref{fig:compare_to_fs}. Our method preserves the inner face features from the identity image, with similar quality to previous methods. However, we also preserve the head shape and hair from the identity image, regardless of that from the attributes image, overall preserving the identity better. Furthermore, we produce results of the same quality even when input images are completely different, as opposed to face swapping methods that are limited to operate on relatively similar images. When input images are different, noticeable artifacts like phantom hair (rows 1,2) and two jaw lines (row 2) may appear. Even when no artifacts are created, the output may be not recognized as either of the input identities and often create an unrealistic face simply because it depicts a very unusual appearance (rows 3, 4).

\label{sec:compare_to_fs}

\begin{table}
    \begin{center}
        \caption{Quantitative Comparison of our method with LORD \cite{gabbay2019demystifying} and FSGAN \cite{nirkin2019fsgan}}
        \label{tab:Quantitative}
        \begin{tabular}{l c c c c}
        \toprule
         \multicolumn{1}{c}{Method}  &  \small{FID}  $\downarrow$ & \small{Identity}  $\uparrow$ & \small{Expression}   $\downarrow$ & \small{Pose}  $\downarrow$ \\
        \midrule
        \small{LORD } & 23.08 & 0.20 $\pm$ 0.11 & 0.085 $\pm$ 0.018 & 13.34 $\pm$ 15.00  \\
        \small{FSGAN} & 8.90 & 0.35 $\pm$ 0.19 & \textbf{0.013 $\pm$ 0.013} & \textbf{6.73 $\pm$ 13.62}  \\
        \small{Ours} & \textbf{4.28} & \textbf{0.60 $\pm$ 0.09} & 0.017 $\pm$ 0.019 & 9.74 $\pm$ 13.16 \\
        \bottomrule
        \end{tabular}
    \end{center}
\end{table}

We further quantify the aforementioned differences in performance by conducting a quantitative comparison to LORD and FSGAN. We evaluate the methods' ability to disentangle and preserve underlying factors composing the human head from different sources, as well as evaluating image quality quantitatively. The evaluation is performed by randomly sampling 10$K$ pairs of images from FFHQ, that are used as identity and attribute inputs. We then apply all methods to infer output images. Next, we assess the preservation of identity, pose and expression from the respective source image, and the quality of the output image.
The results are displayed in \tabref{tab:Quantitative}.

To test identity preservation, we employ state-of-the-art face recognition network, namely ArcFace \cite{deng2019arcface}, and adopt the cosine similarity metric to compare the identity of $I_{id}$ and $I_{out}$. Note that, ArcFace is completely different than the face recognition network used during training, in terms of both the training set and losses. The accuracy of expression preservation is calculated by Euclidean distance between 2$D$ landmarks of $I_{attr}$ and $I_{out}$, inferred using dlib \cite{dlib09}. We normalize the landmarks values by dividing them by the output image resolution, to make the methods comparable. Similarly, pose preservation is measured by the Euclidean distance between Euler angles of $I_{attr}$ and $I_{out}$. For each of the above, we calculate the mean and standard deviation across the test set. Last, we evaluate the quality of the image, specifically on the face region, which is the interest of this work. We sample new 10$K$ images to serve as real images. We then detect \cite{zhang2016joint} and crop faces from the output images and the real images, creating the test sets. Afterward, we calculate the FID \cite{heusel2017gans} score for all methods.
As can be seen in \tabref{tab:Quantitative}, our approach is superior to both methods in terms of image quality and identity preservation, while being comparable to them in expression and pose.

\subsection{Comparison with Reconstruction Methods}
\begin{figure}[ht!]
	\centering
	\includegraphics[width=\linewidth]{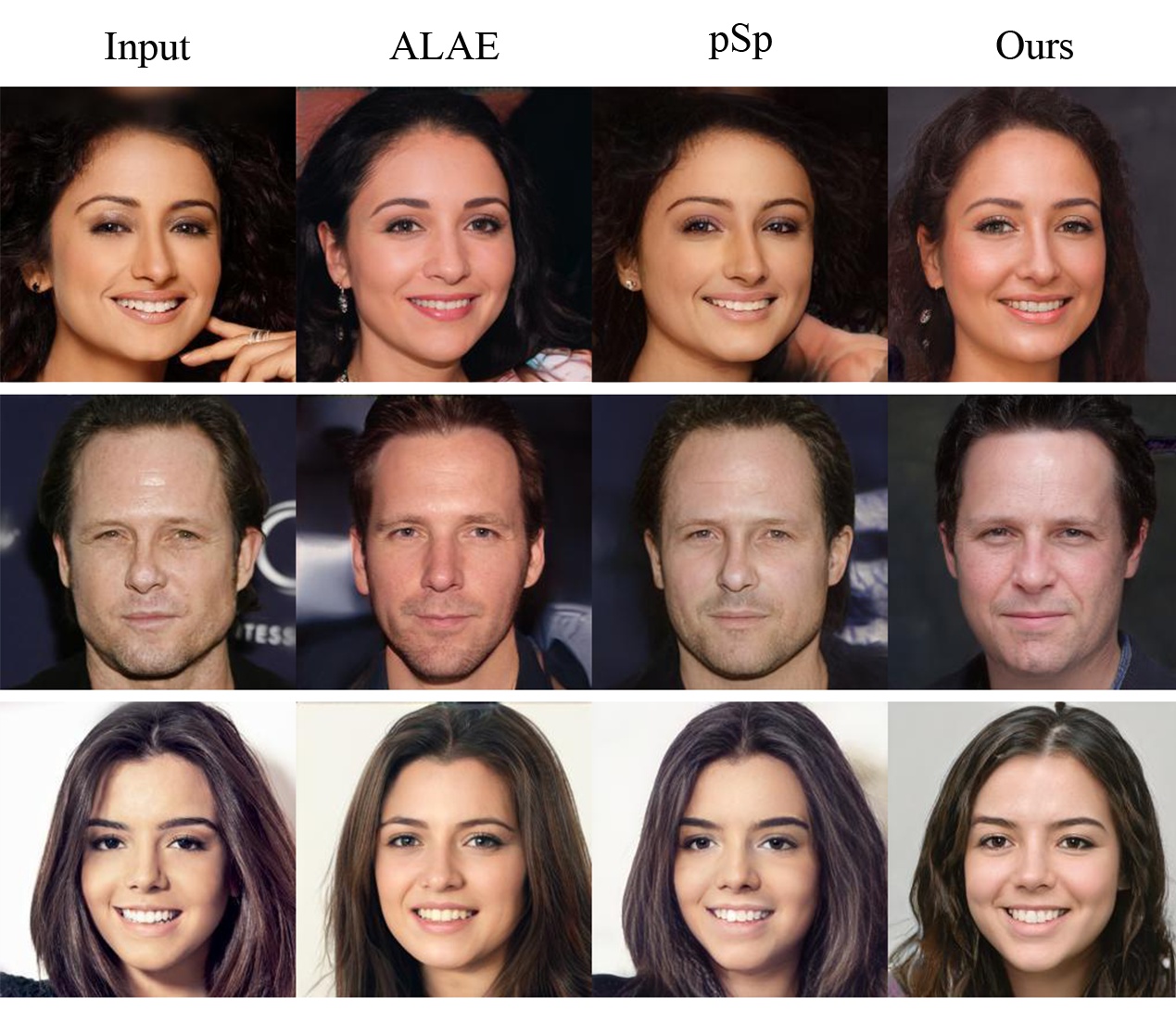}
	\caption{Qualitative comparison to state-of-the-art learning-based reconstruction methods ALAE \cite{pidhorskyi2020adversarial} and pSp \cite{richardson2020encoding}.}
	\label{fig:compare_reconstruction}
\end{figure}

As previously discussed, our task and setting is significantly different than that of \textit{GAN Inversion}. However, if the identity and attribute images fed to our network are the same, our network is implicitly tasked with an inversion problem. For StyleGAN generated faces our inversion is accurate as demonstrated in the diagonal in \Cref{fig:table_results_1,fig:3x3_table_results}. We also study the quality of our reconstruction on real faces by comparing it to two recent state-of-the-art encoder-based inversion methods: ALAE \cite{pidhorskyi2020adversarial} and pSp \cite{richardson2020encoding}. All methods were trained on FFHQ and evaluated on CelebA-HQ \cite{karras2017progressive}.
We perform a quantitative evaluation of the three methods, to assess their reconstruction performance. We randomly sample 5K images from CelebA-HQ for evaluation and evaluate the reconstruction performance by measuring pixel-wise reconstruction with RMSE and PSNR as well as the preservation of semantic features that is identity, expression and pose, in the same manner as performed in \tabref{tab:Quantitative}. Results are displayed in \tabref{tab:reconstruction_table} and a sample of visual results are displayed in \figref{fig:compare_reconstruction}. As can be observed, pSp is superior, while our method is comparable to ALAE.
It is important to stress again, that unlike ALAE and pSp, our method does not aim to reconstruct pixel-level information, as it was devised for the task of disentanglement rather than reconstruction. 
However, it is evident that direct inversion approaches that map directly to the enlarged latent space $\mathcal{W}+$ \cite{richardson2020encoding, abdal2019image2stylegan} perform better. 
This suggests that there is a potential room for improvement in our algorithm by mapping the images into $\mathcal{W}+$.

\label{sec:compare_reconstruction}

\begin{table}
    \begin{center}
        \caption{Reconstruction quantitative comparison with ALAE \cite{pidhorskyi2020adversarial} and pSp \cite{richardson2020encoding} on CelebA-HQ \cite{karras2017progressive}}
        \label{tab:reconstruction_table}
        \begin{tabular}{l c c c c c c}
        \toprule
         \multicolumn{1}{c}{Method} & \small{RMSE}  $\downarrow$ & \small{PSNR}  $\uparrow$ & \small{ID} $\uparrow$ & \small{Expression} $\downarrow$ & \small{Pose} $\downarrow$ \\
        \midrule
        \small{ALAE} & 0.192 & 14.903 & 0.300 & 0.014 & 6.919 \\
        \small{pSp} & \textbf{0.095} & \textbf{21.001} & \textbf{0.821} & \textbf{0.008} & \textbf{4.531} \\
        \small{Ours} & 0.202 & 14.155 & 0.600 & 0.013 & 6.943 \\
        \bottomrule
        \end{tabular}
    \end{center}
\end{table}

\subsection{Ablation Study}
\begin{figure}
	\centering
	\includegraphics[width=0.9\linewidth]{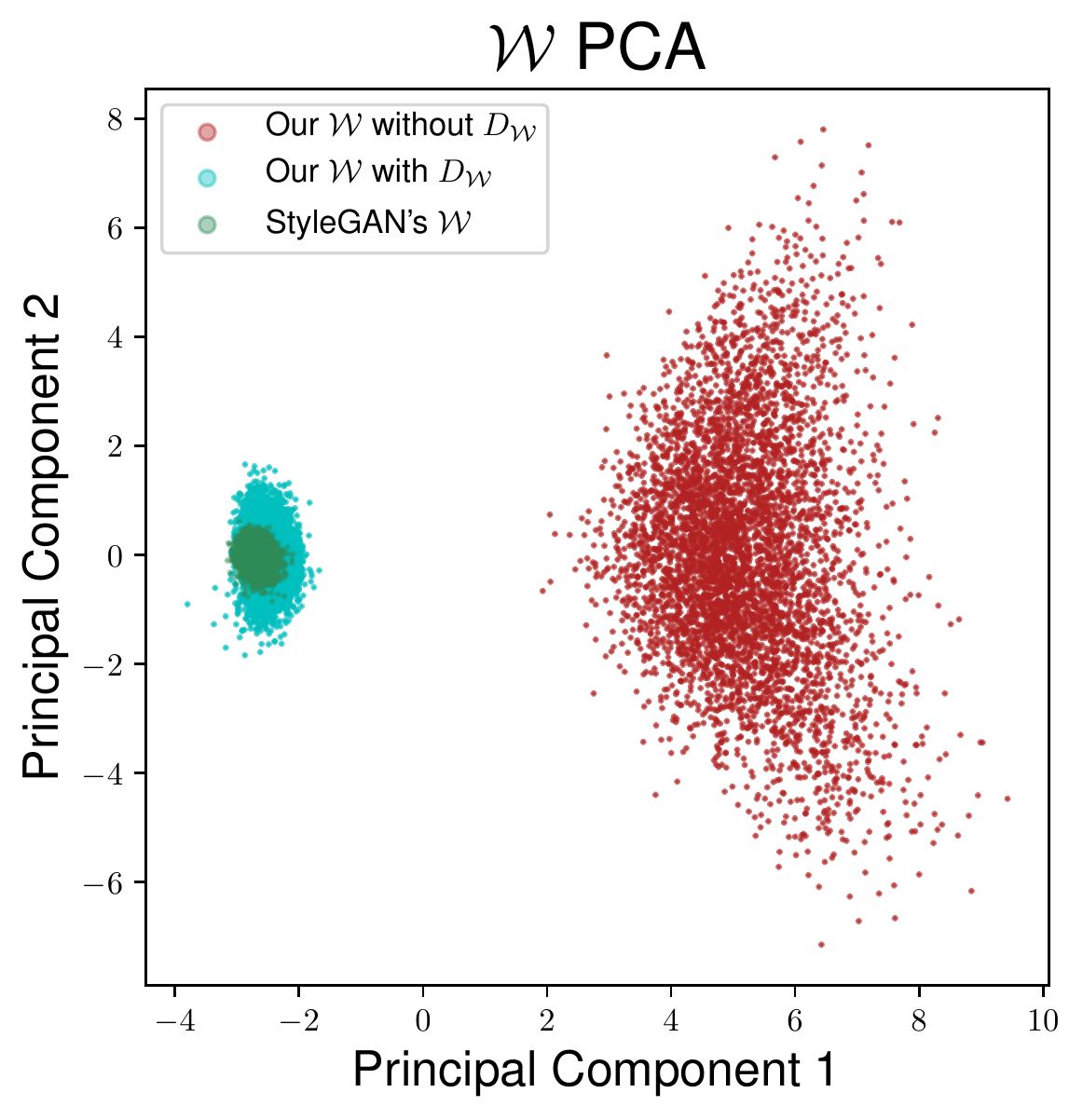}
	\caption{Dimension reduction of different configurations of $\mathcal{W}$, using PCA (n=2). As can be clearly seen, our predicted $\mathcal{W}$ space coincides with StyleGAN's, expanding it only slightly. However, without $D_{\mathcal{W}}$ the predicted $\mathcal{W}$ space is significantly different, stressing the necessity of $D_{\mathcal{W}}$. }
	\label{fig:ablation_W_PCA}
\end{figure}

\begin{figure}
    \centering
    \includegraphics[width=\linewidth]{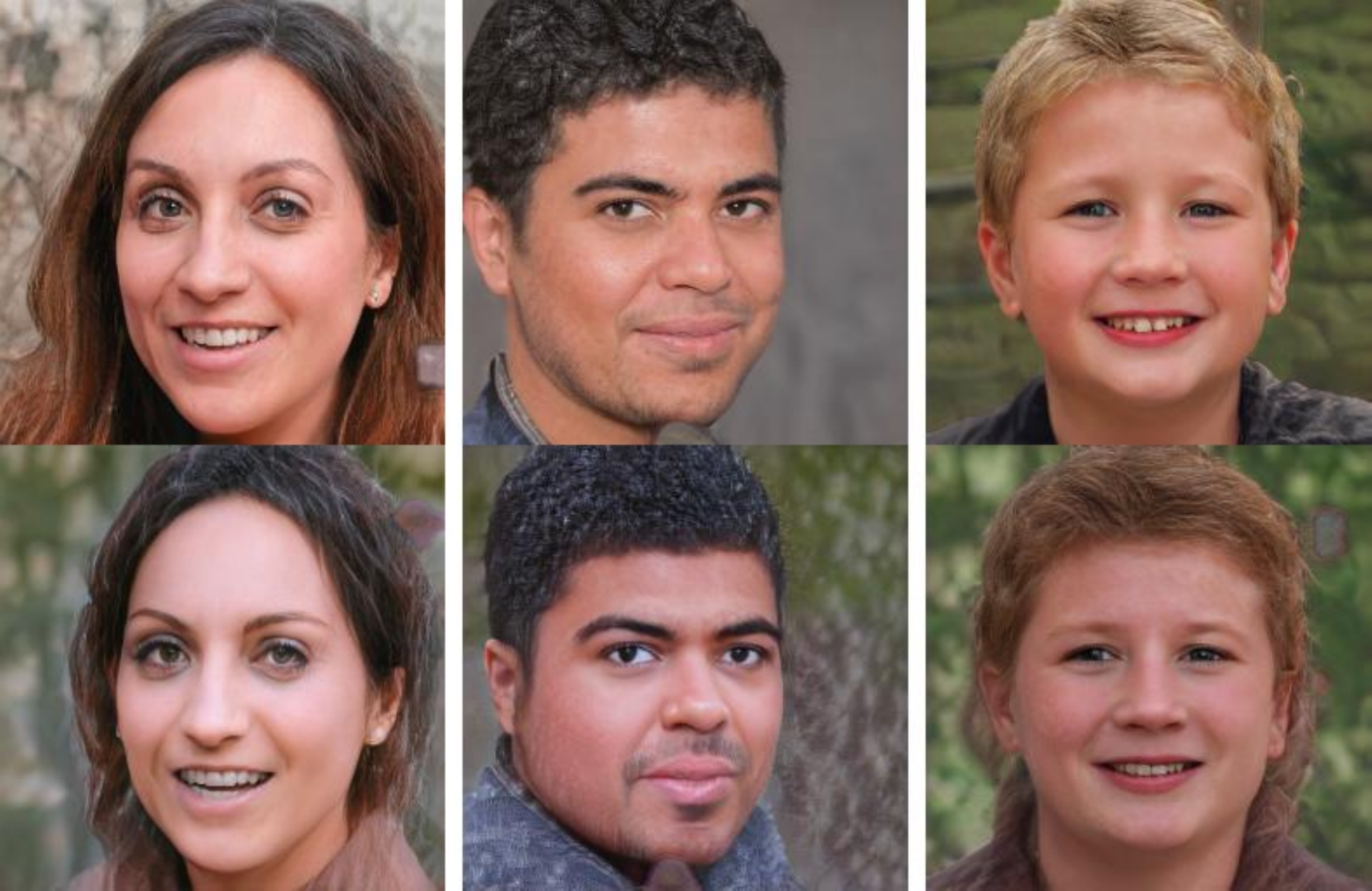}
    \caption{The effect of $D_\mathcal{W}$. Images in the top row were generated by our method. Images in the bottom row were generated by our method, but without $D_{\mathcal{W}}$, on the same inputs. As can be seen, $D_{\mathcal{W}}$ significantly improves image quality, which is also expressed by the FID score that is improved from 6.96 to 4.28.}
    \label{fig:compare_ablation_images}
\end{figure}

In the following we present an ablation study that analyzes the contribution of individual components and losses in our proposed method. 
One of the most intriguing components of our method is $D_{\mathcal{W}}$, which encourages the mapping network to predict latent representations that lie within StyleGAN's $\mathcal{W}$ space. To validate its effect, we train a baseline model, that consists of our architecture without the discriminator $D_{\mathcal{W}}$ and its associated adversarial losses.
We explore and compare three different $\mathcal{W}$ spaces: StyleGAN's original, ours, and the baseline predicted spaces. We sample 10$K$ vectors from each space and compute PCA, visualized in \figref{fig:ablation_W_PCA}. As can be clearly seen, our predicted $\mathcal{W}$ coincides with StyleGAN's $\mathcal{W}$, while the baseline's $\mathcal{W}$ is significantly different, indicating that $D_{\mathcal{W}}$ is crucial for mapping $z$ into the actual $\mathcal{W}$ space of the pre-trained generator.
Furthermore, as stated by Karras et al. \shortcite{karras2019style}, peripheral regions in the latent space often correspond to generated images with inferior quality and resemblance to the real data distribution. Without $D_{\mathcal{W}}$, the predictions may lie in such peripheral regions of $\mathcal{W}$, causing the generation of erroneous images, such as those in \figref{fig:compare_ablation_images}.

Next, we evaluate the contribution of our landmarks consistency loss. This loss is calculated using an accessible off-the-shelf network, pre-trained on an independent dataset and task. In the following, we train our network without $E_{lnd}$, forming a baseline, and show that the supervision of the landmarks network serves merely to improve the attributes preservation, but is not required for disentanglement.
Quantitative evaluation is presented in \tabref{tab:quantitative_ablation}. As can be seen, removing $E_{lnd}$ decreases only the expression and pose preservation. Yet, even with this decrease the baseline is overall superior to LORD in those aspects.
Note that, since this landmarks supervision is not necessary to perform the disentanglement, our overall supervision is significantly weaker than the common class-supervision used by other methods. 

\begin{table}
    \begin{center}
        \caption{Quantitative analysis of $\mathcal{L}_{lnd}$}
        \label{tab:quantitative_ablation}
        \begin{tabular}{l c c c}
        \toprule
         \multicolumn{1}{c}{Method} & \small{Identity}  $\uparrow$ & \small{Expression}   $\downarrow$ & \small{Pose}  $\downarrow$ \\
        \midrule
        \small{LORD} & 0.20 $\pm$ 0.11 & 0.085 $\pm$ 0.018 & 13.34 $\pm$ 15.00  \\
        \small{Ours w/o $\mathcal{L}_{lnd}$} & \textbf{0.63 $\pm$ 0.08} & 0.023 $\pm$ 0.017 & 13.90 $\pm$ 13.36  \\
        \small{Ours} & 0.60 $\pm$ 0.09 & \textbf{0.017 $\pm$ 0.019} & \textbf{9.74 $\pm$ 13.16} \\
        \bottomrule
        \end{tabular}
    \end{center}
\end{table}

\begin{figure*}[h!]
  \centering
  \includegraphics[width=\linewidth]{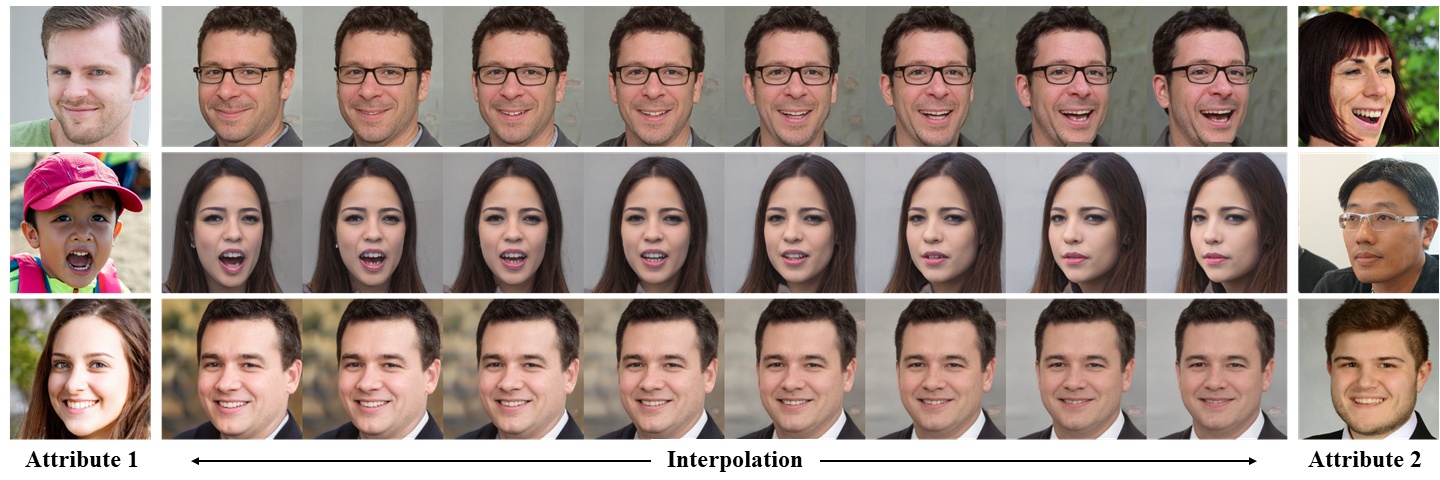}
  \caption{Disentangled interpolation of attributes while preserving identity. In each line, attributes are extracted from two images (both ends of the spectrum), and the identity is extracted from a third image (not shown). For each of these we infer a $w$ vector, and interpolate between the resulting two (middle). }
  \label{fig:interpolate_attr}
  
  \vspace{0.5cm}

  \includegraphics[width=\linewidth]{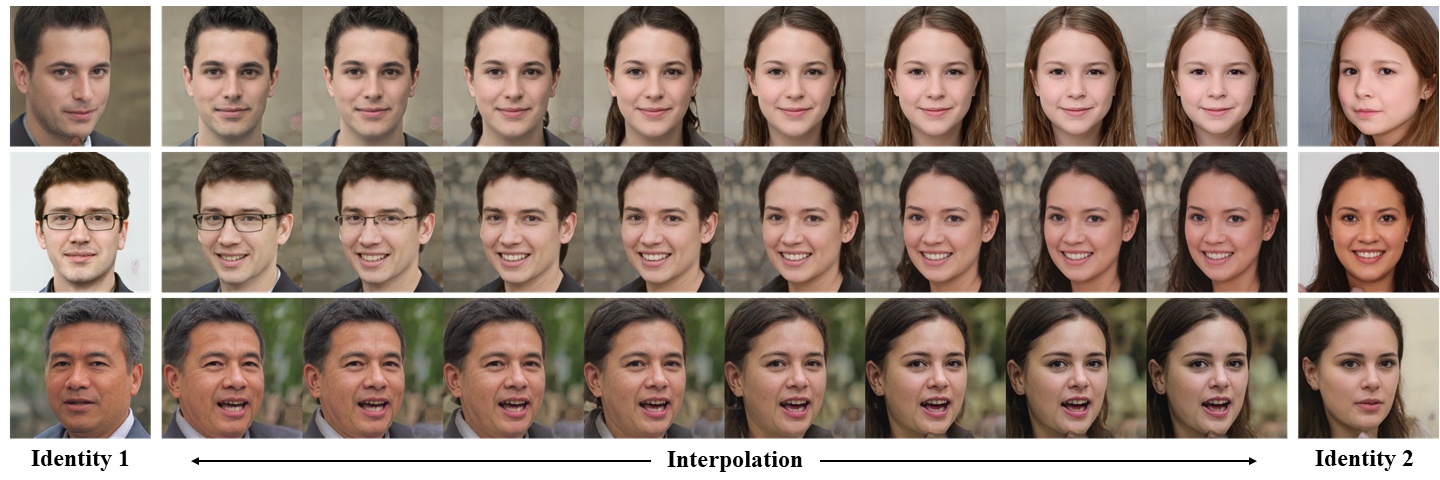}
  \caption{Disentangled interpolation of identity while preserving the other attributes. The setting is identical to the one in \figref{fig:interpolate_attr}, only here the attributes are extracted from the same image (not shown), and the identity is extracted from two images (both ends of the spectrum), and is interpolated in the space of $\mathcal{W}$ (middle).}
   \label{fig:interpolate_id}
   
\end{figure*}

\subsection{Applications}
\label{subsec:applications}
As previously mentioned, our method inherits the properties of the chosen generator $G$ and its latent space. For the running example of this paper, this generator is StyleGAN. Shen et al. \shortcite{shen2019interpreting} have demonstrated that StyleGAN's latent space $\mathcal{W}$ is well behaved, permitting the smooth editing of features via interpolation of the latent code. However, this and other previous art \cite{zhu2020domain} have demonstrated this property for one-dimensional properties, such as age or the extent of a smile. In contrast, our proposed mapping identifies latent codes which represent more involved differences, namely the much discussed high-dimensional identity property, or a combination of expression, pose, and lighting.

In \Cref{fig:interpolate_id,fig:interpolate_attr} we demonstrate the smooth editing of these elaborate features by interpolation of the latent codes, thus showing that the StyleGAN's latent space $\mathcal{W}$ is well behaved even with respect to such features, and that we indeed inherit these advantages. 

In each figure, the interpolated feature is extracted from the images on the far ends, while the constant feature is extracted from a third image that is not shown. From these inputs we infer two $w$ vectors and interpolate between them. The images generated by the interpolated values appear in the middle of each of these figures, and portray a pleasant and natural transition between the various explored properties. 
In \figref{fig:interpolate_attr}, we demonstrate that we accurately and consistently preserve the identity while smoothly and properly interpolate expression, pose and illumination.
Specifically, note the successful interpolation of illumination presented on the bottom line of the figure. In \figref{fig:interpolate_id} we demonstrate that we accurately and consistently preserve the attributes while smoothly interpolating the identity. Note that all images generated during interpolation are of high visual quality and realism.

\begin{figure*}[t]
	\centering
	\includegraphics[width=\linewidth]{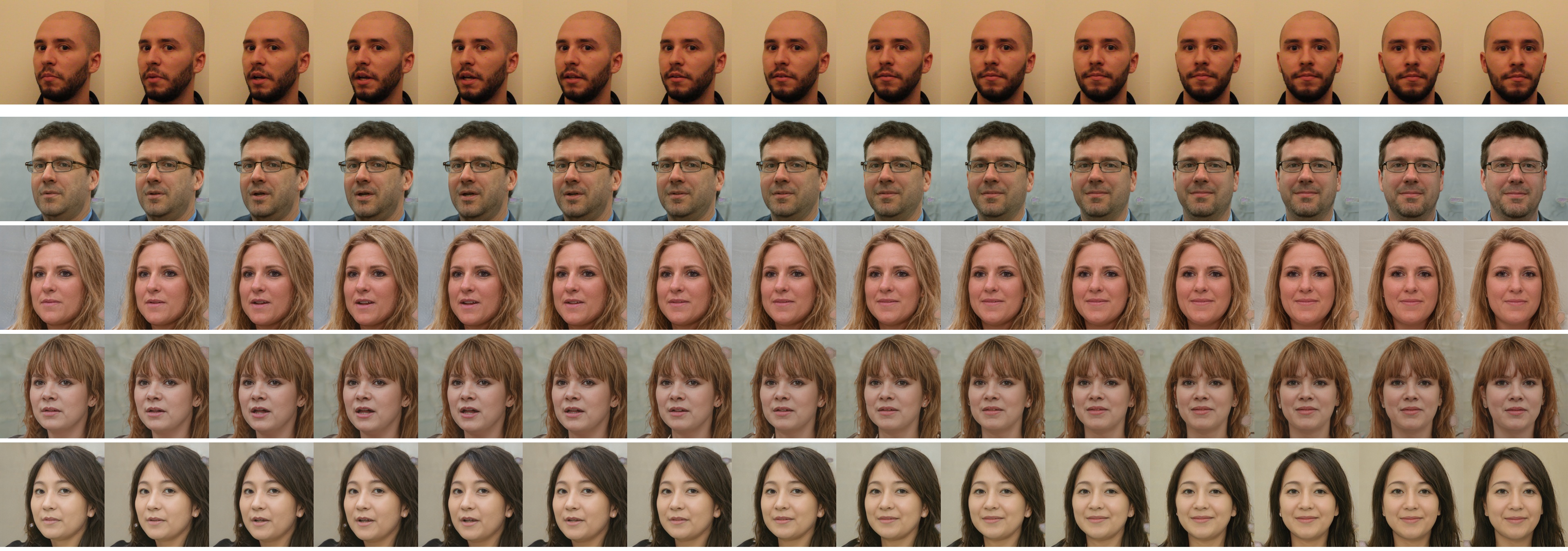}
	\caption{Talking head sequence. The first row consists of consecutive frames from a driving attributes sequence, while the rest of the rows are frames generated by our method. We demonstrate smooth control over facial expression and pose while maintaining constant identity.}
	\label{fig:video_pose_and_talk}

    \vspace{0.5cm}

	\centering
	\includegraphics[width=\linewidth]{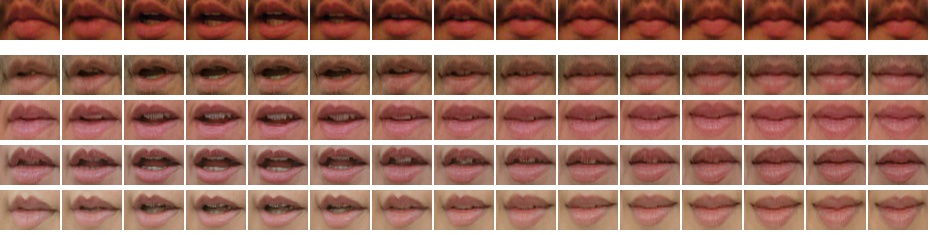}
	\caption{Cropped mouth regions from \figref{fig:video_pose_and_talk}. Our method is able to preserve subtle lips movement, critical for talking head sequence realism.}
	\label{fig:video_only_mouths}
\end{figure*}

Next, we turn to examining the coherency of the network in terms of identity preservation. In other words,  
we examine the stability of the generated identity while perturbing the other attributes. We do this through the generation of sequences. To generate the sequence, we use a single, unseen image to define the target identity, and a sequence of a facial performance to define the rest of the attributes. Generating such a sequence can be considered as a step in the practical direction of the case where the de-identification of the person in the driving input sequence is desired, or when one would like to reenact a single, unseen, given image.
For the case of de-identification, our method is unique because it completely hides the original identity, both from state-of-the-art face recognition networks and from human eyes. This is in contrast to previous methods \cite{gafni2019live}, that perform minimal facial modification to fool face recognition methods. In this case, if the input and output images were put side by side, they would still be recognized as the same person by a human. Our method generates a different person, having a completely different appearance. 
A sample of our results is displayed in \figref{fig:video_pose_and_talk}. Consecutive frames from the driving sequence are displayed in the first row, and the rest of the rows are our results.
Note the different overall appearance of our results compared to the driving sequence. For example, the bottom three rows are generations of different women, all with long hair, while the input is a bald man with a beard. More results, including an animated sequence, can be found in the supplementary material.
As can be seen, this simple approach achieves smooth temporal control over pose and expression, and a very stable and coherent identity for the entire sequence, even though every image was generated completely independently. To better observe details, we crop the mouth region from these frames and display them in \figref{fig:video_only_mouths}. Note the subtle changes in mouth shapes along the sequence.

\label{sec:video_pose_and_talk}

\section{Discussion and Conclusions}
\label{sec:discussion}

This paper presented a novel disentanglement method, applied to the highly challenging domain of human heads. The key idea of mapping the disentangled representation to the latent space of a pre-trained GAN is both novel and crucial. It enables state-of-the-art quality synthesis, while  requiring modest supervision. Through extensive experimentation, we have further demonstrated the effectiveness and versatility of the method. 
We have proposed a novel concept of disentanglement, achieved by mapping to the semantically rich latent space of a pre-trained GAN. This concept is generic and could possibly be applied to other data domains and GAN architectures, assuming its latent space is well behaved. Thus, our work is orthogonal to ongoing research on unconditional image generation, from which the proposed framework can only gain.

Furthermore, this concept concentrates on the preservation of image properties, rather than explicitly reducing "leakage" of information between the two parts of the disentangled representation $\mathcal{Z}$. Preventing leakage is a sufficient condition for disentanglement, but not a necessary one. We separate the latent space $\mathcal{Z}$ in which a disentangled representation is formed and the latent space of the generator $\mathcal{W}$, allowing greater degree of freedom for both tasks. The mapping process connects these two latent spaces, by taking
only the relevant information from each of the parts, and disregards any impurities in the separation of the representation. 

As many works \cite{shen2019interpreting, jahanian2019steerability, zhu2020domain, harkonen2020ganspace} have shown, the latent space of GANs is well-behaved and allows great controlled editing opportunities. All of them, however, have found generic linear directions, along which linear properties can be increased or decreased, in a disentangled fashion. 
Identity, on the other hand, is a complex and high-dimensional factor that cannot be edited by one-dimensional value changes. By exhibiting control over the identity, our method continues the direction of demonstrating the incredible strength and possibilities hidden in the latent space of GANs, and StyleGAN in particular. 

Our approach further relies on the existence of a network, or any other derivable method, to classify, or evaluate, the feature of interest $f$. In the case of human faces, we have also leaned on a similar network to help with identifying facial landmark positions. This was needed due to the extremely sensitive human perception of faces. 
Other than these networks, our dataset does not contain any labeling, nor several images of the same identity. This setting poses a rather weak form of supervision, especially when compared to the common setting of class-supervised disentanglement. 
The supervision in our method is manifested solely through the used pre-trained networks. These, however, can be trained on fundamentally different datasets and tasks, so they do not impose an inhibiting requirement.

As discussed, we inherit the generative capabilities of the used pre-trained generator. Of course, alongside these, we also inherit any limitations the generator might have, including those imposed by its training dataset. For example, the preprocessing method used by StyleGAN aligns heads such that there is no roll angle, and renders yaw rotations to be highly correlated with translation. Thus, StyleGAN-based generations, including ours, inherit these properties. Furthermore, it was recently shown that StyleGAN does not cover the entire manifold of human faces and heads, forcing many approaches \cite{abdal2019image2stylegan, zhu2020domain, pbaylies-stylegan-encoder} to work with an artificially enlarged latent space, named $\mathcal{W}+$. Introducing manipulations on $\mathcal{W}+$ to our methods may significantly increase the expressiveness of our model, but may come at the cost of both generation and disentanglement qualities. We leave this investigation as an exciting avenue for future work. Regardless, this work introduces a powerful concept, where generative networks are employed as "backbones" for disentanglement tasks, which can most probably be explored much further in future research.

\begin{acks}
We thank the anonymous reviewers for their comments.
We also thank Tal Hassner, Yuval Nirkin, Aviv Gabbay and Mingchao Sun for their help and useful suggestions. This work was supported by the Israel Science Foundation (grant no. 2366/16 and 2472/17) and in part by the National Science Foundation of China General Program grant No. 61772317
\end{acks}

\bibliographystyle{ACM-Reference-Format}
\bibliography{egbib}


\begin{thebibliography}{74}


\ifx \showCODEN    \undefined \def \showCODEN     #1{\unskip}     \fi
\ifx \showDOI      \undefined \def \showDOI       #1{#1}\fi
\ifx \showISBNx    \undefined \def \showISBNx     #1{\unskip}     \fi
\ifx \showISBNxiii \undefined \def \showISBNxiii  #1{\unskip}     \fi
\ifx \showISSN     \undefined \def \showISSN      #1{\unskip}     \fi
\ifx \showLCCN     \undefined \def \showLCCN      #1{\unskip}     \fi
\ifx \shownote     \undefined \def \shownote      #1{#1}          \fi
\ifx \showarticletitle \undefined \def \showarticletitle #1{#1}   \fi
\ifx \showURL      \undefined \def \showURL       {\relax}        \fi
\providecommand\bibfield[2]{#2}
\providecommand\bibinfo[2]{#2}
\providecommand\natexlab[1]{#1}
\providecommand\showeprint[2][]{arXiv:#2}

\bibitem[\protect\citeauthoryear{Abdal, Qin, and Wonka}{Abdal
  et~al\mbox{.}}{2019}]%
        {abdal2019image2stylegan}
\bibfield{author}{\bibinfo{person}{Rameen Abdal}, \bibinfo{person}{Yipeng Qin},
  {and} \bibinfo{person}{Peter Wonka}.} \bibinfo{year}{2019}\natexlab{}.
\newblock \showarticletitle{Image2StyleGAN: How to Embed Images Into the
  StyleGAN Latent Space?}. In \bibinfo{booktitle}{\emph{Proceedings of the IEEE
  International Conference on Computer Vision}}. \bibinfo{pages}{4432--4441}.
\newblock


\bibitem[\protect\citeauthoryear{Aberman, Wu, Lischinski, Chen, and
  Cohen-Or}{Aberman et~al\mbox{.}}{2019}]%
        {aberman2019learning}
\bibfield{author}{\bibinfo{person}{Kfir Aberman}, \bibinfo{person}{Rundi Wu},
  \bibinfo{person}{Dani Lischinski}, \bibinfo{person}{Baoquan Chen}, {and}
  \bibinfo{person}{Daniel Cohen-Or}.} \bibinfo{year}{2019}\natexlab{}.
\newblock \showarticletitle{Learning character-agnostic motion for motion
  retargeting in 2D}.
\newblock \bibinfo{journal}{\emph{arXiv preprint arXiv:1905.01680}}
  (\bibinfo{year}{2019}).
\newblock


\bibitem[\protect\citeauthoryear{Abudarham, Shkiller, and Yovel}{Abudarham
  et~al\mbox{.}}{2019}]%
        {abudarham2019critical}
\bibfield{author}{\bibinfo{person}{Naphtali Abudarham}, \bibinfo{person}{Lior
  Shkiller}, {and} \bibinfo{person}{Galit Yovel}.}
  \bibinfo{year}{2019}\natexlab{}.
\newblock \showarticletitle{Critical features for face recognition}.
\newblock \bibinfo{journal}{\emph{Cognition}}  \bibinfo{volume}{182}
  (\bibinfo{year}{2019}), \bibinfo{pages}{73--83}.
\newblock


\bibitem[\protect\citeauthoryear{Averbuch-Elor, Cohen-Or, Kopf, and
  Cohen}{Averbuch-Elor et~al\mbox{.}}{2017}]%
        {elor2017bringingPortraits}
\bibfield{author}{\bibinfo{person}{Hadar Averbuch-Elor},
  \bibinfo{person}{Daniel Cohen-Or}, \bibinfo{person}{Johannes Kopf}, {and}
  \bibinfo{person}{Michael~F. Cohen}.} \bibinfo{year}{2017}\natexlab{}.
\newblock \showarticletitle{Bringing Portraits to Life}.
\newblock \bibinfo{journal}{\emph{ACM Transactions on Graphics (Proceeding of
  SIGGRAPH Asia 2017)}} \bibinfo{volume}{36}, \bibinfo{number}{6}
  (\bibinfo{year}{2017}), \bibinfo{pages}{196}.
\newblock


\bibitem[\protect\citeauthoryear{Bao, Chen, Wen, Li, and Hua}{Bao
  et~al\mbox{.}}{2018}]%
        {bao2018towards}
\bibfield{author}{\bibinfo{person}{Jianmin Bao}, \bibinfo{person}{Dong Chen},
  \bibinfo{person}{Fang Wen}, \bibinfo{person}{Houqiang Li}, {and}
  \bibinfo{person}{Gang Hua}.} \bibinfo{year}{2018}\natexlab{}.
\newblock \showarticletitle{Towards open-set identity preserving face
  synthesis}. In \bibinfo{booktitle}{\emph{Proceedings of the IEEE Conference
  on Computer Vision and Pattern Recognition}}. \bibinfo{pages}{6713--6722}.
\newblock


\bibitem[\protect\citeauthoryear{{Baylies}}{{Baylies}}{2019}]%
        {pbaylies-stylegan-encoder}
\bibfield{author}{\bibinfo{person}{{Baylies}}.}
  \bibinfo{year}{2019}\natexlab{}.
\newblock \bibinfo{title}{stylegan-encoder}.
\newblock
  \bibinfo{howpublished}{\url{https://github.com/pbaylies/stylegan-encoder}}.
\newblock
\newblock
\shownote{Accessed: April 2020.}


\bibitem[\protect\citeauthoryear{Bengio, Courville, and Vincent}{Bengio
  et~al\mbox{.}}{2013}]%
        {bengio2013representation}
\bibfield{author}{\bibinfo{person}{Yoshua Bengio}, \bibinfo{person}{Aaron
  Courville}, {and} \bibinfo{person}{Pascal Vincent}.}
  \bibinfo{year}{2013}\natexlab{}.
\newblock \showarticletitle{Representation learning: A review and new
  perspectives}.
\newblock \bibinfo{journal}{\emph{IEEE transactions on pattern analysis and
  machine intelligence}} \bibinfo{volume}{35}, \bibinfo{number}{8}
  (\bibinfo{year}{2013}), \bibinfo{pages}{1798--1828}.
\newblock


\bibitem[\protect\citeauthoryear{Bouchacourt, Tomioka, and Nowozin}{Bouchacourt
  et~al\mbox{.}}{2018}]%
        {bouchacourt2018multi}
\bibfield{author}{\bibinfo{person}{Diane Bouchacourt}, \bibinfo{person}{Ryota
  Tomioka}, {and} \bibinfo{person}{Sebastian Nowozin}.}
  \bibinfo{year}{2018}\natexlab{}.
\newblock \showarticletitle{Multi-level variational autoencoder: Learning
  disentangled representations from grouped observations}. In
  \bibinfo{booktitle}{\emph{Thirty-Second AAAI Conference on Artificial
  Intelligence}}.
\newblock


\bibitem[\protect\citeauthoryear{Cao, Shen, Xie, Parkhi, and Zisserman}{Cao
  et~al\mbox{.}}{2018}]%
        {cao2018vggface2}
\bibfield{author}{\bibinfo{person}{Qiong Cao}, \bibinfo{person}{Li Shen},
  \bibinfo{person}{Weidi Xie}, \bibinfo{person}{Omkar~M Parkhi}, {and}
  \bibinfo{person}{Andrew Zisserman}.} \bibinfo{year}{2018}\natexlab{}.
\newblock \showarticletitle{Vggface2: A dataset for recognising faces across
  pose and age}. In \bibinfo{booktitle}{\emph{2018 13th IEEE International
  Conference on Automatic Face \& Gesture Recognition (FG 2018)}}. IEEE,
  \bibinfo{pages}{67--74}.
\newblock


\bibitem[\protect\citeauthoryear{Chen, Duan, Houthooft, Schulman, Sutskever,
  and Abbeel}{Chen et~al\mbox{.}}{2016}]%
        {chen2016infogan}
\bibfield{author}{\bibinfo{person}{Xi Chen}, \bibinfo{person}{Yan Duan},
  \bibinfo{person}{Rein Houthooft}, \bibinfo{person}{John Schulman},
  \bibinfo{person}{Ilya Sutskever}, {and} \bibinfo{person}{Pieter Abbeel}.}
  \bibinfo{year}{2016}\natexlab{}.
\newblock \showarticletitle{Infogan: Interpretable representation learning by
  information maximizing generative adversarial nets}. In
  \bibinfo{booktitle}{\emph{Advances in neural information processing
  systems}}. \bibinfo{pages}{2172--2180}.
\newblock


\bibitem[\protect\citeauthoryear{Choi, Choi, Kim, Ha, Kim, and Choo}{Choi
  et~al\mbox{.}}{2018}]%
        {choi2018stargan}
\bibfield{author}{\bibinfo{person}{Yunjey Choi}, \bibinfo{person}{Minje Choi},
  \bibinfo{person}{Munyoung Kim}, \bibinfo{person}{Jung-Woo Ha},
  \bibinfo{person}{Sunghun Kim}, {and} \bibinfo{person}{Jaegul Choo}.}
  \bibinfo{year}{2018}\natexlab{}.
\newblock \showarticletitle{Stargan: Unified generative adversarial networks
  for multi-domain image-to-image translation}. In
  \bibinfo{booktitle}{\emph{Proceedings of the IEEE conference on computer
  vision and pattern recognition}}. \bibinfo{pages}{8789--8797}.
\newblock


\bibitem[\protect\citeauthoryear{Choi, Uh, Yoo, and Ha}{Choi
  et~al\mbox{.}}{2019}]%
        {choi2019stargan}
\bibfield{author}{\bibinfo{person}{Yunjey Choi}, \bibinfo{person}{Youngjung
  Uh}, \bibinfo{person}{Jaejun Yoo}, {and} \bibinfo{person}{Jung-Woo Ha}.}
  \bibinfo{year}{2019}\natexlab{}.
\newblock \showarticletitle{StarGAN v2: Diverse Image Synthesis for Multiple
  Domains}.
\newblock \bibinfo{journal}{\emph{arXiv preprint arXiv:1912.01865}}
  (\bibinfo{year}{2019}).
\newblock


\bibitem[\protect\citeauthoryear{Creswell and Bharath}{Creswell and
  Bharath}{2018}]%
        {creswell2018inverting}
\bibfield{author}{\bibinfo{person}{Antonia Creswell} {and}
  \bibinfo{person}{Anil~Anthony Bharath}.} \bibinfo{year}{2018}\natexlab{}.
\newblock \showarticletitle{Inverting the generator of a generative adversarial
  network}.
\newblock \bibinfo{journal}{\emph{IEEE transactions on neural networks and
  learning systems}} \bibinfo{volume}{30}, \bibinfo{number}{7}
  (\bibinfo{year}{2018}), \bibinfo{pages}{1967--1974}.
\newblock


\bibitem[\protect\citeauthoryear{{deepfakes}}{{deepfakes}}{2019}]%
        {deepfake}
\bibfield{author}{\bibinfo{person}{{deepfakes}}.}
  \bibinfo{year}{2019}\natexlab{}.
\newblock \bibinfo{title}{faceswap}.
\newblock \bibinfo{howpublished}{\url{https://github.com/deepfakes/faceswap}}.
\newblock
\newblock
\shownote{Accessed: April 2020.}


\bibitem[\protect\citeauthoryear{Deng, Guo, Xue, and Zafeiriou}{Deng
  et~al\mbox{.}}{2019}]%
        {deng2019arcface}
\bibfield{author}{\bibinfo{person}{Jiankang Deng}, \bibinfo{person}{Jia Guo},
  \bibinfo{person}{Niannan Xue}, {and} \bibinfo{person}{Stefanos Zafeiriou}.}
  \bibinfo{year}{2019}\natexlab{}.
\newblock \showarticletitle{Arcface: Additive angular margin loss for deep face
  recognition}. In \bibinfo{booktitle}{\emph{Proceedings of the IEEE Conference
  on Computer Vision and Pattern Recognition}}. \bibinfo{pages}{4690--4699}.
\newblock


\bibitem[\protect\citeauthoryear{Denton, Hutchinson, Mitchell, and
  Gebru}{Denton et~al\mbox{.}}{2019}]%
        {denton2019detecting}
\bibfield{author}{\bibinfo{person}{Emily Denton}, \bibinfo{person}{Ben
  Hutchinson}, \bibinfo{person}{Margaret Mitchell}, {and}
  \bibinfo{person}{Timnit Gebru}.} \bibinfo{year}{2019}\natexlab{}.
\newblock \showarticletitle{Detecting bias with generative counterfactual face
  attribute augmentation}.
\newblock \bibinfo{journal}{\emph{arXiv preprint arXiv:1906.06439}}
  (\bibinfo{year}{2019}).
\newblock


\bibitem[\protect\citeauthoryear{Denton et~al\mbox{.}}{Denton
  et~al\mbox{.}}{2017}]%
        {denton2017unsupervised}
\bibfield{author}{\bibinfo{person}{Emily~L Denton} {et~al\mbox{.}}}
  \bibinfo{year}{2017}\natexlab{}.
\newblock \showarticletitle{Unsupervised learning of disentangled
  representations from video}. In \bibinfo{booktitle}{\emph{Advances in neural
  information processing systems}}. \bibinfo{pages}{4414--4423}.
\newblock


\bibitem[\protect\citeauthoryear{Feng, Kittler, Awais, Huber, and Wu}{Feng
  et~al\mbox{.}}{2018}]%
        {feng2018wing}
\bibfield{author}{\bibinfo{person}{Zhen-Hua Feng}, \bibinfo{person}{Josef
  Kittler}, \bibinfo{person}{Muhammad Awais}, \bibinfo{person}{Patrik Huber},
  {and} \bibinfo{person}{Xiao-Jun Wu}.} \bibinfo{year}{2018}\natexlab{}.
\newblock \showarticletitle{Wing loss for robust facial landmark localisation
  with convolutional neural networks}. In \bibinfo{booktitle}{\emph{Proceedings
  of the IEEE Conference on Computer Vision and Pattern Recognition}}.
  \bibinfo{pages}{2235--2245}.
\newblock


\bibitem[\protect\citeauthoryear{Gabbay and Hoshen}{Gabbay and Hoshen}{2019}]%
        {gabbay2019demystifying}
\bibfield{author}{\bibinfo{person}{Aviv Gabbay} {and} \bibinfo{person}{Yedid
  Hoshen}.} \bibinfo{year}{2019}\natexlab{}.
\newblock \showarticletitle{Demystifying Inter-Class Disentanglement}.
\newblock \bibinfo{journal}{\emph{arXiv preprint arXiv:1906.11796}}
  (\bibinfo{year}{2019}).
\newblock


\bibitem[\protect\citeauthoryear{Gafni, Wolf, and Taigman}{Gafni
  et~al\mbox{.}}{2019}]%
        {gafni2019live}
\bibfield{author}{\bibinfo{person}{Oran Gafni}, \bibinfo{person}{Lior Wolf},
  {and} \bibinfo{person}{Yaniv Taigman}.} \bibinfo{year}{2019}\natexlab{}.
\newblock \showarticletitle{Live Face De-Identification in Video}. In
  \bibinfo{booktitle}{\emph{Proceedings of the IEEE International Conference on
  Computer Vision}}. \bibinfo{pages}{9378--9387}.
\newblock


\bibitem[\protect\citeauthoryear{Goetschalckx, Andonian, Oliva, and
  Isola}{Goetschalckx et~al\mbox{.}}{2019}]%
        {goetschalckx2019ganalyze}
\bibfield{author}{\bibinfo{person}{Lore Goetschalckx}, \bibinfo{person}{Alex
  Andonian}, \bibinfo{person}{Aude Oliva}, {and} \bibinfo{person}{Phillip
  Isola}.} \bibinfo{year}{2019}\natexlab{}.
\newblock \bibinfo{title}{GANalyze: Toward Visual Definitions of Cognitive
  Image Properties}.
\newblock
\newblock
\showeprint[arxiv]{cs.CV/1906.10112}


\bibitem[\protect\citeauthoryear{Goodfellow, Pouget-Abadie, Mirza, Xu,
  Warde-Farley, Ozair, Courville, and Bengio}{Goodfellow et~al\mbox{.}}{2014}]%
        {goodfellow2014generative}
\bibfield{author}{\bibinfo{person}{Ian Goodfellow}, \bibinfo{person}{Jean
  Pouget-Abadie}, \bibinfo{person}{Mehdi Mirza}, \bibinfo{person}{Bing Xu},
  \bibinfo{person}{David Warde-Farley}, \bibinfo{person}{Sherjil Ozair},
  \bibinfo{person}{Aaron Courville}, {and} \bibinfo{person}{Yoshua Bengio}.}
  \bibinfo{year}{2014}\natexlab{}.
\newblock \showarticletitle{Generative Adversarial Nets}.
\newblock In \bibinfo{booktitle}{\emph{Advances in Neural Information
  Processing Systems 27}}, \bibfield{editor}{\bibinfo{person}{Z.~Ghahramani},
  \bibinfo{person}{M.~Welling}, \bibinfo{person}{C.~Cortes},
  \bibinfo{person}{N.~D. Lawrence}, {and} \bibinfo{person}{K.~Q. Weinberger}}
  (Eds.). \bibinfo{publisher}{Curran Associates, Inc.},
  \bibinfo{pages}{2672--2680}.
\newblock
\urldef\tempurl%
\url{http://papers.nips.cc/paper/5423-generative-adversarial-nets.pdf}
\showURL{%
\tempurl}


\bibitem[\protect\citeauthoryear{Hadad, Wolf, and Shahar}{Hadad
  et~al\mbox{.}}{2018}]%
        {hadad2018two}
\bibfield{author}{\bibinfo{person}{Naama Hadad}, \bibinfo{person}{Lior Wolf},
  {and} \bibinfo{person}{Moni Shahar}.} \bibinfo{year}{2018}\natexlab{}.
\newblock \showarticletitle{A two-step disentanglement method}. In
  \bibinfo{booktitle}{\emph{Proceedings of the IEEE Conference on Computer
  Vision and Pattern Recognition}}. \bibinfo{pages}{772--780}.
\newblock


\bibitem[\protect\citeauthoryear{H{\"a}rk{\"o}nen, Hertzmann, Lehtinen, and
  Paris}{H{\"a}rk{\"o}nen et~al\mbox{.}}{2020}]%
        {harkonen2020ganspace}
\bibfield{author}{\bibinfo{person}{Erik H{\"a}rk{\"o}nen},
  \bibinfo{person}{Aaron Hertzmann}, \bibinfo{person}{Jaakko Lehtinen}, {and}
  \bibinfo{person}{Sylvain Paris}.} \bibinfo{year}{2020}\natexlab{}.
\newblock \showarticletitle{GANSpace: Discovering Interpretable GAN Controls}.
\newblock \bibinfo{journal}{\emph{arXiv preprint arXiv:2004.02546}}
  (\bibinfo{year}{2020}).
\newblock


\bibitem[\protect\citeauthoryear{He, Zhang, Ren, and Sun}{He
  et~al\mbox{.}}{2015}]%
        {he2015delving}
\bibfield{author}{\bibinfo{person}{Kaiming He}, \bibinfo{person}{Xiangyu
  Zhang}, \bibinfo{person}{Shaoqing Ren}, {and} \bibinfo{person}{Jian Sun}.}
  \bibinfo{year}{2015}\natexlab{}.
\newblock \showarticletitle{Delving deep into rectifiers: Surpassing
  human-level performance on imagenet classification}. In
  \bibinfo{booktitle}{\emph{Proceedings of the IEEE international conference on
  computer vision}}. \bibinfo{pages}{1026--1034}.
\newblock


\bibitem[\protect\citeauthoryear{He, Zhang, Ren, and Sun}{He
  et~al\mbox{.}}{2016}]%
        {he2016deep}
\bibfield{author}{\bibinfo{person}{Kaiming He}, \bibinfo{person}{Xiangyu
  Zhang}, \bibinfo{person}{Shaoqing Ren}, {and} \bibinfo{person}{Jian Sun}.}
  \bibinfo{year}{2016}\natexlab{}.
\newblock \showarticletitle{Deep residual learning for image recognition}. In
  \bibinfo{booktitle}{\emph{Proceedings of the IEEE conference on computer
  vision and pattern recognition}}. \bibinfo{pages}{770--778}.
\newblock


\bibitem[\protect\citeauthoryear{Heusel, Ramsauer, Unterthiner, Nessler, and
  Hochreiter}{Heusel et~al\mbox{.}}{2017}]%
        {heusel2017gans}
\bibfield{author}{\bibinfo{person}{Martin Heusel}, \bibinfo{person}{Hubert
  Ramsauer}, \bibinfo{person}{Thomas Unterthiner}, \bibinfo{person}{Bernhard
  Nessler}, {and} \bibinfo{person}{Sepp Hochreiter}.}
  \bibinfo{year}{2017}\natexlab{}.
\newblock \showarticletitle{Gans trained by a two time-scale update rule
  converge to a local nash equilibrium}. In \bibinfo{booktitle}{\emph{Advances
  in neural information processing systems}}. \bibinfo{pages}{6626--6637}.
\newblock


\bibitem[\protect\citeauthoryear{Higgins, Matthey, Pal, Burgess, Glorot,
  Botvinick, Mohamed, and Lerchner}{Higgins et~al\mbox{.}}{2017}]%
        {higgins2017beta}
\bibfield{author}{\bibinfo{person}{Irina Higgins}, \bibinfo{person}{Loic
  Matthey}, \bibinfo{person}{Arka Pal}, \bibinfo{person}{Christopher Burgess},
  \bibinfo{person}{Xavier Glorot}, \bibinfo{person}{Matthew Botvinick},
  \bibinfo{person}{Shakir Mohamed}, {and} \bibinfo{person}{Alexander
  Lerchner}.} \bibinfo{year}{2017}\natexlab{}.
\newblock \showarticletitle{beta-VAE: Learning Basic Visual Concepts with a
  Constrained Variational Framework.}
\newblock \bibinfo{journal}{\emph{Iclr}} \bibinfo{volume}{2},
  \bibinfo{number}{5} (\bibinfo{year}{2017}), \bibinfo{pages}{6}.
\newblock


\bibitem[\protect\citeauthoryear{Huang, He, Sun, Tan, et~al\mbox{.}}{Huang
  et~al\mbox{.}}{2018}]%
        {huang2018introvae}
\bibfield{author}{\bibinfo{person}{Huaibo Huang}, \bibinfo{person}{Ran He},
  \bibinfo{person}{Zhenan Sun}, \bibinfo{person}{Tieniu Tan}, {et~al\mbox{.}}}
  \bibinfo{year}{2018}\natexlab{}.
\newblock \showarticletitle{Introvae: Introspective variational autoencoders
  for photographic image synthesis}. In \bibinfo{booktitle}{\emph{Advances in
  neural information processing systems}}. \bibinfo{pages}{52--63}.
\newblock


\bibitem[\protect\citeauthoryear{Huang and Belongie}{Huang and
  Belongie}{2017}]%
        {huang2017arbitrary}
\bibfield{author}{\bibinfo{person}{Xun Huang} {and} \bibinfo{person}{Serge
  Belongie}.} \bibinfo{year}{2017}\natexlab{}.
\newblock \showarticletitle{Arbitrary style transfer in real-time with adaptive
  instance normalization}. In \bibinfo{booktitle}{\emph{Proceedings of the IEEE
  International Conference on Computer Vision}}. \bibinfo{pages}{1501--1510}.
\newblock


\bibitem[\protect\citeauthoryear{Jahanian, Chai, and Isola}{Jahanian
  et~al\mbox{.}}{2019}]%
        {jahanian2019steerability}
\bibfield{author}{\bibinfo{person}{Ali Jahanian}, \bibinfo{person}{Lucy Chai},
  {and} \bibinfo{person}{Phillip Isola}.} \bibinfo{year}{2019}\natexlab{}.
\newblock \showarticletitle{On the "steerability" of generative adversarial
  networks}.
\newblock \bibinfo{journal}{\emph{arXiv preprint arXiv:1907.07171}}
  (\bibinfo{year}{2019}).
\newblock


\bibitem[\protect\citeauthoryear{Johnson, Alahi, and Fei-Fei}{Johnson
  et~al\mbox{.}}{2016}]%
        {johnson2016perceptual}
\bibfield{author}{\bibinfo{person}{Justin Johnson}, \bibinfo{person}{Alexandre
  Alahi}, {and} \bibinfo{person}{Li Fei-Fei}.} \bibinfo{year}{2016}\natexlab{}.
\newblock \showarticletitle{Perceptual losses for real-time style transfer and
  super-resolution}. In \bibinfo{booktitle}{\emph{European conference on
  computer vision}}. Springer, \bibinfo{pages}{694--711}.
\newblock


\bibitem[\protect\citeauthoryear{Karras, Aila, Laine, and Lehtinen}{Karras
  et~al\mbox{.}}{2017}]%
        {karras2017progressive}
\bibfield{author}{\bibinfo{person}{Tero Karras}, \bibinfo{person}{Timo Aila},
  \bibinfo{person}{Samuli Laine}, {and} \bibinfo{person}{Jaakko Lehtinen}.}
  \bibinfo{year}{2017}\natexlab{}.
\newblock \showarticletitle{Progressive growing of gans for improved quality,
  stability, and variation}.
\newblock \bibinfo{journal}{\emph{arXiv preprint arXiv:1710.10196}}
  (\bibinfo{year}{2017}).
\newblock


\bibitem[\protect\citeauthoryear{Karras, Laine, and Aila}{Karras
  et~al\mbox{.}}{2019b}]%
        {karras2019style}
\bibfield{author}{\bibinfo{person}{Tero Karras}, \bibinfo{person}{Samuli
  Laine}, {and} \bibinfo{person}{Timo Aila}.} \bibinfo{year}{2019}\natexlab{b}.
\newblock \showarticletitle{A style-based generator architecture for generative
  adversarial networks}. In \bibinfo{booktitle}{\emph{Proceedings of the IEEE
  Conference on Computer Vision and Pattern Recognition}}.
  \bibinfo{pages}{4401--4410}.
\newblock


\bibitem[\protect\citeauthoryear{Karras, Laine, Aittala, Hellsten, Lehtinen,
  and Aila}{Karras et~al\mbox{.}}{2019a}]%
        {karras2019analyzing}
\bibfield{author}{\bibinfo{person}{Tero Karras}, \bibinfo{person}{Samuli
  Laine}, \bibinfo{person}{Miika Aittala}, \bibinfo{person}{Janne Hellsten},
  \bibinfo{person}{Jaakko Lehtinen}, {and} \bibinfo{person}{Timo Aila}.}
  \bibinfo{year}{2019}\natexlab{a}.
\newblock \showarticletitle{Analyzing and improving the image quality of
  stylegan}.
\newblock \bibinfo{journal}{\emph{arXiv preprint arXiv:1912.04958}}
  (\bibinfo{year}{2019}).
\newblock


\bibitem[\protect\citeauthoryear{Kim, Garrido, Tewari, Xu, Thies, Nie{\ss}ner,
  P{\'e}rez, Richardt, Zollh{\"o}fer, and Theobalt}{Kim et~al\mbox{.}}{2018}]%
        {kim2018deep}
\bibfield{author}{\bibinfo{person}{Hyeongwoo Kim}, \bibinfo{person}{Pablo
  Garrido}, \bibinfo{person}{Ayush Tewari}, \bibinfo{person}{Weipeng Xu},
  \bibinfo{person}{Justus Thies}, \bibinfo{person}{Matthias Nie{\ss}ner},
  \bibinfo{person}{Patrick P{\'e}rez}, \bibinfo{person}{Christian Richardt},
  \bibinfo{person}{Michael Zollh{\"o}fer}, {and} \bibinfo{person}{Christian
  Theobalt}.} \bibinfo{year}{2018}\natexlab{}.
\newblock \showarticletitle{Deep video portraits}.
\newblock \bibinfo{journal}{\emph{ACM Transactions on Graphics (TOG)}}
  \bibinfo{volume}{37}, \bibinfo{number}{4} (\bibinfo{year}{2018}),
  \bibinfo{pages}{1--14}.
\newblock


\bibitem[\protect\citeauthoryear{Kim and Mnih}{Kim and Mnih}{2018}]%
        {kim2018disentangling}
\bibfield{author}{\bibinfo{person}{Hyunjik Kim} {and} \bibinfo{person}{Andriy
  Mnih}.} \bibinfo{year}{2018}\natexlab{}.
\newblock \showarticletitle{Disentangling by factorising}.
\newblock \bibinfo{journal}{\emph{arXiv preprint arXiv:1802.05983}}
  (\bibinfo{year}{2018}).
\newblock


\bibitem[\protect\citeauthoryear{King}{King}{2009}]%
        {dlib09}
\bibfield{author}{\bibinfo{person}{Davis~E. King}.}
  \bibinfo{year}{2009}\natexlab{}.
\newblock \showarticletitle{Dlib-ml: A Machine Learning Toolkit}.
\newblock \bibinfo{journal}{\emph{Journal of Machine Learning Research}}
  \bibinfo{volume}{10} (\bibinfo{year}{2009}), \bibinfo{pages}{1755--1758}.
\newblock


\bibitem[\protect\citeauthoryear{Kingma and Ba}{Kingma and Ba}{2015}]%
        {DBLP:journals/corr/KingmaB14}
\bibfield{author}{\bibinfo{person}{Diederik~P. Kingma} {and}
  \bibinfo{person}{Jimmy Ba}.} \bibinfo{year}{2015}\natexlab{}.
\newblock \showarticletitle{Adam: {A} Method for Stochastic Optimization}. In
  \bibinfo{booktitle}{\emph{3rd International Conference on Learning
  Representations, {ICLR} 2015, San Diego, CA, USA, May 7-9, 2015, Conference
  Track Proceedings}}, \bibfield{editor}{\bibinfo{person}{Yoshua Bengio} {and}
  \bibinfo{person}{Yann LeCun}} (Eds.).
\newblock
\urldef\tempurl%
\url{http://arxiv.org/abs/1412.6980}
\showURL{%
\tempurl}


\bibitem[\protect\citeauthoryear{Li, Bao, Yang, Chen, and Wen}{Li
  et~al\mbox{.}}{2019}]%
        {li2019faceshifter}
\bibfield{author}{\bibinfo{person}{Lingzhi Li}, \bibinfo{person}{Jianmin Bao},
  \bibinfo{person}{Hao Yang}, \bibinfo{person}{Dong Chen}, {and}
  \bibinfo{person}{Fang Wen}.} \bibinfo{year}{2019}\natexlab{}.
\newblock \showarticletitle{FaceShifter: Towards High Fidelity And Occlusion
  Aware Face Swapping}.
\newblock \bibinfo{journal}{\emph{arXiv preprint arXiv:1912.13457}}
  (\bibinfo{year}{2019}).
\newblock


\bibitem[\protect\citeauthoryear{Lipton and Tripathi}{Lipton and
  Tripathi}{2017}]%
        {lipton2017precise}
\bibfield{author}{\bibinfo{person}{Zachary~C Lipton} {and}
  \bibinfo{person}{Subarna Tripathi}.} \bibinfo{year}{2017}\natexlab{}.
\newblock \showarticletitle{Precise recovery of latent vectors from generative
  adversarial networks}.
\newblock \bibinfo{journal}{\emph{arXiv preprint arXiv:1702.04782}}
  (\bibinfo{year}{2017}).
\newblock


\bibitem[\protect\citeauthoryear{Liu, Breuel, and Kautz}{Liu
  et~al\mbox{.}}{2017}]%
        {liu2017unsupervised}
\bibfield{author}{\bibinfo{person}{Ming-Yu Liu}, \bibinfo{person}{Thomas
  Breuel}, {and} \bibinfo{person}{Jan Kautz}.} \bibinfo{year}{2017}\natexlab{}.
\newblock \showarticletitle{Unsupervised image-to-image translation networks}.
  In \bibinfo{booktitle}{\emph{Advances in neural information processing
  systems}}. \bibinfo{pages}{700--708}.
\newblock


\bibitem[\protect\citeauthoryear{Liu, Wei, Shao, Sheng, Yan, and Wang}{Liu
  et~al\mbox{.}}{2018}]%
        {liu2018exploring}
\bibfield{author}{\bibinfo{person}{Yu Liu}, \bibinfo{person}{Fangyin Wei},
  \bibinfo{person}{Jing Shao}, \bibinfo{person}{Lu Sheng},
  \bibinfo{person}{Junjie Yan}, {and} \bibinfo{person}{Xiaogang Wang}.}
  \bibinfo{year}{2018}\natexlab{}.
\newblock \showarticletitle{Exploring disentangled feature representation
  beyond face identification}. In \bibinfo{booktitle}{\emph{Proceedings of the
  IEEE Conference on Computer Vision and Pattern Recognition}}.
  \bibinfo{pages}{2080--2089}.
\newblock


\bibitem[\protect\citeauthoryear{Liu, Luo, Wang, and Tang}{Liu
  et~al\mbox{.}}{2015}]%
        {liu2015faceattributes}
\bibfield{author}{\bibinfo{person}{Ziwei Liu}, \bibinfo{person}{Ping Luo},
  \bibinfo{person}{Xiaogang Wang}, {and} \bibinfo{person}{Xiaoou Tang}.}
  \bibinfo{year}{2015}\natexlab{}.
\newblock \showarticletitle{Deep Learning Face Attributes in the Wild}. In
  \bibinfo{booktitle}{\emph{Proceedings of International Conference on Computer
  Vision (ICCV)}}.
\newblock


\bibitem[\protect\citeauthoryear{Locatello, Bauer, Lucic, R{\"a}tsch, Gelly,
  Sch{\"o}lkopf, and Bachem}{Locatello et~al\mbox{.}}{2018}]%
        {locatello2018challenging}
\bibfield{author}{\bibinfo{person}{Francesco Locatello},
  \bibinfo{person}{Stefan Bauer}, \bibinfo{person}{Mario Lucic},
  \bibinfo{person}{Gunnar R{\"a}tsch}, \bibinfo{person}{Sylvain Gelly},
  \bibinfo{person}{Bernhard Sch{\"o}lkopf}, {and} \bibinfo{person}{Olivier
  Bachem}.} \bibinfo{year}{2018}\natexlab{}.
\newblock \showarticletitle{Challenging common assumptions in the unsupervised
  learning of disentangled representations}.
\newblock \bibinfo{journal}{\emph{arXiv preprint arXiv:1811.12359}}
  (\bibinfo{year}{2018}).
\newblock


\bibitem[\protect\citeauthoryear{Luo, Xu, Tang, and Lv}{Luo
  et~al\mbox{.}}{2017}]%
        {luo2017learning}
\bibfield{author}{\bibinfo{person}{Junyu Luo}, \bibinfo{person}{Yong Xu},
  \bibinfo{person}{Chenwei Tang}, {and} \bibinfo{person}{Jiancheng Lv}.}
  \bibinfo{year}{2017}\natexlab{}.
\newblock \showarticletitle{Learning inverse mapping by autoencoder based
  generative adversarial nets}. In \bibinfo{booktitle}{\emph{International
  Conference on Neural Information Processing}}. Springer,
  \bibinfo{pages}{207--216}.
\newblock


\bibitem[\protect\citeauthoryear{Mathieu, Rainforth, Siddharth, and
  Teh}{Mathieu et~al\mbox{.}}{2018}]%
        {mathieu2018disentangling}
\bibfield{author}{\bibinfo{person}{Emile Mathieu}, \bibinfo{person}{Tom
  Rainforth}, \bibinfo{person}{N. Siddharth}, {and} \bibinfo{person}{Yee~Whye
  Teh}.} \bibinfo{year}{2018}\natexlab{}.
\newblock \bibinfo{title}{Disentangling Disentanglement in Variational
  Autoencoders}.
\newblock
\newblock
\showeprint[arxiv]{stat.ML/1812.02833}


\bibitem[\protect\citeauthoryear{Mescheder, Geiger, and Nowozin}{Mescheder
  et~al\mbox{.}}{2018}]%
        {mescheder2018training}
\bibfield{author}{\bibinfo{person}{Lars Mescheder}, \bibinfo{person}{Andreas
  Geiger}, {and} \bibinfo{person}{Sebastian Nowozin}.}
  \bibinfo{year}{2018}\natexlab{}.
\newblock \showarticletitle{Which training methods for GANs do actually
  converge?}
\newblock \bibinfo{journal}{\emph{arXiv preprint arXiv:1801.04406}}
  (\bibinfo{year}{2018}).
\newblock


\bibitem[\protect\citeauthoryear{Mori et~al\mbox{.}}{Mori
  et~al\mbox{.}}{1970}]%
        {mori1970uncanny}
\bibfield{author}{\bibinfo{person}{Masahiro Mori} {et~al\mbox{.}}}
  \bibinfo{year}{1970}\natexlab{}.
\newblock \showarticletitle{The uncanny valley}.
\newblock \bibinfo{journal}{\emph{Energy}} \bibinfo{volume}{7},
  \bibinfo{number}{4} (\bibinfo{year}{1970}), \bibinfo{pages}{33--35}.
\newblock


\bibitem[\protect\citeauthoryear{Nirkin, Keller, and Hassner}{Nirkin
  et~al\mbox{.}}{2019}]%
        {nirkin2019fsgan}
\bibfield{author}{\bibinfo{person}{Yuval Nirkin}, \bibinfo{person}{Yosi
  Keller}, {and} \bibinfo{person}{Tal Hassner}.}
  \bibinfo{year}{2019}\natexlab{}.
\newblock \showarticletitle{Fsgan: Subject agnostic face swapping and
  reenactment}. In \bibinfo{booktitle}{\emph{Proceedings of the IEEE
  International Conference on Computer Vision}}. \bibinfo{pages}{7184--7193}.
\newblock


\bibitem[\protect\citeauthoryear{Perarnau, Van De~Weijer, Raducanu, and
  {\'A}lvarez}{Perarnau et~al\mbox{.}}{2016}]%
        {perarnau2016invertible}
\bibfield{author}{\bibinfo{person}{Guim Perarnau}, \bibinfo{person}{Joost Van
  De~Weijer}, \bibinfo{person}{Bogdan Raducanu}, {and} \bibinfo{person}{Jose~M
  {\'A}lvarez}.} \bibinfo{year}{2016}\natexlab{}.
\newblock \showarticletitle{Invertible conditional gans for image editing}.
\newblock \bibinfo{journal}{\emph{arXiv preprint arXiv:1611.06355}}
  (\bibinfo{year}{2016}).
\newblock


\bibitem[\protect\citeauthoryear{Pidhorskyi, Adjeroh, and Doretto}{Pidhorskyi
  et~al\mbox{.}}{2020}]%
        {pidhorskyi2020adversarial}
\bibfield{author}{\bibinfo{person}{Stanislav Pidhorskyi},
  \bibinfo{person}{Donald~A Adjeroh}, {and} \bibinfo{person}{Gianfranco
  Doretto}.} \bibinfo{year}{2020}\natexlab{}.
\newblock \showarticletitle{Adversarial Latent Autoencoders}. In
  \bibinfo{booktitle}{\emph{Proceedings of the IEEE Computer Society Conference
  on Computer Vision and Pattern Recognition (CVPR)}}.
\newblock
\newblock
\shownote{[to appear].}


\bibitem[\protect\citeauthoryear{Pumarola, Agudo, Martinez, Sanfeliu, and
  Moreno-Noguer}{Pumarola et~al\mbox{.}}{2018}]%
        {pumarola2018ganimation}
\bibfield{author}{\bibinfo{person}{Albert Pumarola}, \bibinfo{person}{Antonio
  Agudo}, \bibinfo{person}{Aleix~M Martinez}, \bibinfo{person}{Alberto
  Sanfeliu}, {and} \bibinfo{person}{Francesc Moreno-Noguer}.}
  \bibinfo{year}{2018}\natexlab{}.
\newblock \showarticletitle{Ganimation: Anatomically-aware facial animation
  from a single image}. In \bibinfo{booktitle}{\emph{Proceedings of the
  European Conference on Computer Vision (ECCV)}}. \bibinfo{pages}{818--833}.
\newblock


\bibitem[\protect\citeauthoryear{Reed, Zhang, Zhang, and Lee}{Reed
  et~al\mbox{.}}{2015}]%
        {reed2015deep}
\bibfield{author}{\bibinfo{person}{Scott~E Reed}, \bibinfo{person}{Yi Zhang},
  \bibinfo{person}{Yuting Zhang}, {and} \bibinfo{person}{Honglak Lee}.}
  \bibinfo{year}{2015}\natexlab{}.
\newblock \showarticletitle{Deep visual analogy-making}. In
  \bibinfo{booktitle}{\emph{Advances in neural information processing
  systems}}. \bibinfo{pages}{1252--1260}.
\newblock


\bibitem[\protect\citeauthoryear{Richardson, Alaluf, Patashnik, Nitzan, Azar,
  Shapiro, and Cohen-Or}{Richardson et~al\mbox{.}}{2020}]%
        {richardson2020encoding}
\bibfield{author}{\bibinfo{person}{Elad Richardson}, \bibinfo{person}{Yuval
  Alaluf}, \bibinfo{person}{Or Patashnik}, \bibinfo{person}{Yotam Nitzan},
  \bibinfo{person}{Yaniv Azar}, \bibinfo{person}{Stav Shapiro}, {and}
  \bibinfo{person}{Daniel Cohen-Or}.} \bibinfo{year}{2020}\natexlab{}.
\newblock \showarticletitle{Encoding in Style: a StyleGAN Encoder for
  Image-to-Image Translation}.
\newblock \bibinfo{journal}{\emph{arXiv preprint arXiv:2008.00951}}
  (\bibinfo{year}{2020}).
\newblock


\bibitem[\protect\citeauthoryear{Sendik, Lischinski, and Cohen-Or}{Sendik
  et~al\mbox{.}}{2019}]%
        {sendik2019s}
\bibfield{author}{\bibinfo{person}{Omry Sendik}, \bibinfo{person}{Dani
  Lischinski}, {and} \bibinfo{person}{Daniel Cohen-Or}.}
  \bibinfo{year}{2019}\natexlab{}.
\newblock \showarticletitle{What's in a Face? Metric Learning for Face
  Characterization}. In \bibinfo{booktitle}{\emph{Computer Graphics Forum}},
  Vol.~\bibinfo{volume}{38}. Wiley Online Library, \bibinfo{pages}{405--416}.
\newblock


\bibitem[\protect\citeauthoryear{Shen, Gu, Tang, and Zhou}{Shen
  et~al\mbox{.}}{2019}]%
        {shen2019interpreting}
\bibfield{author}{\bibinfo{person}{Yujun Shen}, \bibinfo{person}{Jinjin Gu},
  \bibinfo{person}{Xiaoou Tang}, {and} \bibinfo{person}{Bolei Zhou}.}
  \bibinfo{year}{2019}\natexlab{}.
\newblock \showarticletitle{Interpreting the latent space of gans for semantic
  face editing}.
\newblock \bibinfo{journal}{\emph{arXiv preprint arXiv:1907.10786}}
  (\bibinfo{year}{2019}).
\newblock


\bibitem[\protect\citeauthoryear{Shen, Luo, Yan, Wang, and Tang}{Shen
  et~al\mbox{.}}{2018}]%
        {shen2018faceid}
\bibfield{author}{\bibinfo{person}{Yujun Shen}, \bibinfo{person}{Ping Luo},
  \bibinfo{person}{Junjie Yan}, \bibinfo{person}{Xiaogang Wang}, {and}
  \bibinfo{person}{Xiaoou Tang}.} \bibinfo{year}{2018}\natexlab{}.
\newblock \showarticletitle{Faceid-gan: Learning a symmetry three-player gan
  for identity-preserving face synthesis}. In
  \bibinfo{booktitle}{\emph{Proceedings of the IEEE Conference on Computer
  Vision and Pattern Recognition}}. \bibinfo{pages}{821--830}.
\newblock


\bibitem[\protect\citeauthoryear{Sinha and Poggio}{Sinha and Poggio}{1996}]%
        {sinha1996think}
\bibfield{author}{\bibinfo{person}{Pawan Sinha} {and} \bibinfo{person}{Tomaso
  Poggio}.} \bibinfo{year}{1996}\natexlab{}.
\newblock \showarticletitle{I think I know that face...}
\newblock \bibinfo{journal}{\emph{Nature}} \bibinfo{volume}{384},
  \bibinfo{number}{6608} (\bibinfo{year}{1996}), \bibinfo{pages}{404--404}.
\newblock


\bibitem[\protect\citeauthoryear{Sun, Tewari, Xu, Fritz, Theobalt, and
  Schiele}{Sun et~al\mbox{.}}{2018}]%
        {sun2018hybrid}
\bibfield{author}{\bibinfo{person}{Qianru Sun}, \bibinfo{person}{Ayush Tewari},
  \bibinfo{person}{Weipeng Xu}, \bibinfo{person}{Mario Fritz},
  \bibinfo{person}{Christian Theobalt}, {and} \bibinfo{person}{Bernt Schiele}.}
  \bibinfo{year}{2018}\natexlab{}.
\newblock \showarticletitle{A hybrid model for identity obfuscation by face
  replacement}. In \bibinfo{booktitle}{\emph{Proceedings of the European
  Conference on Computer Vision (ECCV)}}. \bibinfo{pages}{553--569}.
\newblock


\bibitem[\protect\citeauthoryear{Szegedy, Vanhoucke, Ioffe, Shlens, and
  Wojna}{Szegedy et~al\mbox{.}}{2016}]%
        {szegedy2016rethinking}
\bibfield{author}{\bibinfo{person}{Christian Szegedy}, \bibinfo{person}{Vincent
  Vanhoucke}, \bibinfo{person}{Sergey Ioffe}, \bibinfo{person}{Jon Shlens},
  {and} \bibinfo{person}{Zbigniew Wojna}.} \bibinfo{year}{2016}\natexlab{}.
\newblock \showarticletitle{Rethinking the inception architecture for computer
  vision}. In \bibinfo{booktitle}{\emph{Proceedings of the IEEE conference on
  computer vision and pattern recognition}}. \bibinfo{pages}{2818--2826}.
\newblock


\bibitem[\protect\citeauthoryear{Tewari, Elgharib, Bharaj, Bernard, Seidel,
  P{\'e}rez, Zollh{\"o}fer, and Theobalt}{Tewari et~al\mbox{.}}{2020}]%
        {tewari2020stylerig}
\bibfield{author}{\bibinfo{person}{Ayush Tewari}, \bibinfo{person}{Mohamed
  Elgharib}, \bibinfo{person}{Gaurav Bharaj}, \bibinfo{person}{Florian
  Bernard}, \bibinfo{person}{Hans-Peter Seidel}, \bibinfo{person}{Patrick
  P{\'e}rez}, \bibinfo{person}{Michael Zollh{\"o}fer}, {and}
  \bibinfo{person}{Christian Theobalt}.} \bibinfo{year}{2020}\natexlab{}.
\newblock \showarticletitle{StyleRig: Rigging StyleGAN for 3D Control over
  Portrait Images}.
\newblock \bibinfo{journal}{\emph{arXiv preprint arXiv:2004.00121}}
  (\bibinfo{year}{2020}).
\newblock


\bibitem[\protect\citeauthoryear{Thies, Zollhofer, Stamminger, Theobalt, and
  Nie{\ss}ner}{Thies et~al\mbox{.}}{2016}]%
        {thies2016face2face}
\bibfield{author}{\bibinfo{person}{Justus Thies}, \bibinfo{person}{Michael
  Zollhofer}, \bibinfo{person}{Marc Stamminger}, \bibinfo{person}{Christian
  Theobalt}, {and} \bibinfo{person}{Matthias Nie{\ss}ner}.}
  \bibinfo{year}{2016}\natexlab{}.
\newblock \showarticletitle{Face2face: Real-time face capture and reenactment
  of rgb videos}. In \bibinfo{booktitle}{\emph{Proceedings of the IEEE
  conference on computer vision and pattern recognition}}.
  \bibinfo{pages}{2387--2395}.
\newblock


\bibitem[\protect\citeauthoryear{Thies, Zollh{\"o}fer, Theobalt, Stamminger,
  and Nie{\ss}ner}{Thies et~al\mbox{.}}{2018}]%
        {thies2018headon}
\bibfield{author}{\bibinfo{person}{Justus Thies}, \bibinfo{person}{Michael
  Zollh{\"o}fer}, \bibinfo{person}{Christian Theobalt}, \bibinfo{person}{Marc
  Stamminger}, {and} \bibinfo{person}{Matthias Nie{\ss}ner}.}
  \bibinfo{year}{2018}\natexlab{}.
\newblock \showarticletitle{Headon: Real-time reenactment of human portrait
  videos}.
\newblock \bibinfo{journal}{\emph{ACM Transactions on Graphics (TOG)}}
  \bibinfo{volume}{37}, \bibinfo{number}{4} (\bibinfo{year}{2018}),
  \bibinfo{pages}{1--13}.
\newblock


\bibitem[\protect\citeauthoryear{Tian, Peng, Zhao, Zhang, and Metaxas}{Tian
  et~al\mbox{.}}{2018}]%
        {tian2018cr}
\bibfield{author}{\bibinfo{person}{Yu Tian}, \bibinfo{person}{Xi Peng},
  \bibinfo{person}{Long Zhao}, \bibinfo{person}{Shaoting Zhang}, {and}
  \bibinfo{person}{Dimitris~N Metaxas}.} \bibinfo{year}{2018}\natexlab{}.
\newblock \showarticletitle{CR-GAN: learning complete representations for
  multi-view generation}.
\newblock \bibinfo{journal}{\emph{arXiv preprint arXiv:1806.11191}}
  (\bibinfo{year}{2018}).
\newblock


\bibitem[\protect\citeauthoryear{Toseeb, Keeble, and Bryant}{Toseeb
  et~al\mbox{.}}{2012}]%
        {toseeb2012significance}
\bibfield{author}{\bibinfo{person}{Umar Toseeb}, \bibinfo{person}{David~RT
  Keeble}, {and} \bibinfo{person}{Eleanor~J Bryant}.}
  \bibinfo{year}{2012}\natexlab{}.
\newblock \showarticletitle{The significance of hair for face recognition}.
\newblock \bibinfo{journal}{\emph{PloS one}} \bibinfo{volume}{7},
  \bibinfo{number}{3} (\bibinfo{year}{2012}).
\newblock


\bibitem[\protect\citeauthoryear{Tschannen, Bachem, and Lucic}{Tschannen
  et~al\mbox{.}}{2018}]%
        {tschannen2018recent}
\bibfield{author}{\bibinfo{person}{Michael Tschannen}, \bibinfo{person}{Olivier
  Bachem}, {and} \bibinfo{person}{Mario Lucic}.}
  \bibinfo{year}{2018}\natexlab{}.
\newblock \showarticletitle{Recent advances in autoencoder-based representation
  learning}.
\newblock \bibinfo{journal}{\emph{arXiv preprint arXiv:1812.05069}}
  (\bibinfo{year}{2018}).
\newblock


\bibitem[\protect\citeauthoryear{Wu, Yang, and Ling}{Wu et~al\mbox{.}}{2018}]%
        {wu2018privacy}
\bibfield{author}{\bibinfo{person}{Yifan Wu}, \bibinfo{person}{Fan Yang}, {and}
  \bibinfo{person}{Haibin Ling}.} \bibinfo{year}{2018}\natexlab{}.
\newblock \showarticletitle{Privacy-protective-gan for face de-identification}.
\newblock \bibinfo{journal}{\emph{arXiv preprint arXiv:1806.08906}}
  (\bibinfo{year}{2018}).
\newblock


\bibitem[\protect\citeauthoryear{Xiao, Liu, and Lee}{Xiao
  et~al\mbox{.}}{2019}]%
        {xiao2019identity}
\bibfield{author}{\bibinfo{person}{Fanyi Xiao}, \bibinfo{person}{Haotian Liu},
  {and} \bibinfo{person}{Yong~Jae Lee}.} \bibinfo{year}{2019}\natexlab{}.
\newblock \showarticletitle{Identity from here, Pose from there:
  Self-supervised Disentanglement and Generation of Objects using Unlabeled
  Videos}. In \bibinfo{booktitle}{\emph{Proceedings of the IEEE International
  Conference on Computer Vision}}. \bibinfo{pages}{7013--7022}.
\newblock


\bibitem[\protect\citeauthoryear{Zakharov, Shysheya, Burkov, and
  Lempitsky}{Zakharov et~al\mbox{.}}{2019}]%
        {zakharov2019few}
\bibfield{author}{\bibinfo{person}{Egor Zakharov}, \bibinfo{person}{Aliaksandra
  Shysheya}, \bibinfo{person}{Egor Burkov}, {and} \bibinfo{person}{Victor
  Lempitsky}.} \bibinfo{year}{2019}\natexlab{}.
\newblock \showarticletitle{Few-shot adversarial learning of realistic neural
  talking head models}. In \bibinfo{booktitle}{\emph{Proceedings of the IEEE
  International Conference on Computer Vision}}. \bibinfo{pages}{9459--9468}.
\newblock


\bibitem[\protect\citeauthoryear{Zhang, Zhang, Li, and Qiao}{Zhang
  et~al\mbox{.}}{2016}]%
        {zhang2016joint}
\bibfield{author}{\bibinfo{person}{Kaipeng Zhang}, \bibinfo{person}{Zhanpeng
  Zhang}, \bibinfo{person}{Zhifeng Li}, {and} \bibinfo{person}{Yu Qiao}.}
  \bibinfo{year}{2016}\natexlab{}.
\newblock \showarticletitle{Joint face detection and alignment using multitask
  cascaded convolutional networks}.
\newblock \bibinfo{journal}{\emph{IEEE Signal Processing Letters}}
  \bibinfo{volume}{23}, \bibinfo{number}{10} (\bibinfo{year}{2016}),
  \bibinfo{pages}{1499--1503}.
\newblock


\bibitem[\protect\citeauthoryear{Zhao, Gallo, Frosio, and Kautz}{Zhao
  et~al\mbox{.}}{2016}]%
        {zhao2016loss}
\bibfield{author}{\bibinfo{person}{Hang Zhao}, \bibinfo{person}{Orazio Gallo},
  \bibinfo{person}{Iuri Frosio}, {and} \bibinfo{person}{Jan Kautz}.}
  \bibinfo{year}{2016}\natexlab{}.
\newblock \showarticletitle{Loss functions for image restoration with neural
  networks}.
\newblock \bibinfo{journal}{\emph{IEEE Transactions on computational imaging}}
  \bibinfo{volume}{3}, \bibinfo{number}{1} (\bibinfo{year}{2016}),
  \bibinfo{pages}{47--57}.
\newblock


\bibitem[\protect\citeauthoryear{Zhu, Shen, Zhao, and Zhou}{Zhu
  et~al\mbox{.}}{2020}]%
        {zhu2020domain}
\bibfield{author}{\bibinfo{person}{Jiapeng Zhu}, \bibinfo{person}{Yujun Shen},
  \bibinfo{person}{Deli Zhao}, {and} \bibinfo{person}{Bolei Zhou}.}
  \bibinfo{year}{2020}\natexlab{}.
\newblock \showarticletitle{In-Domain GAN Inversion for Real Image Editing}.
\newblock \bibinfo{journal}{\emph{arXiv preprint arXiv:2004.00049}}
  (\bibinfo{year}{2020}).
\newblock


\bibitem[\protect\citeauthoryear{Zhu, Kr{\"a}henb{\"u}hl, Shechtman, and
  Efros}{Zhu et~al\mbox{.}}{2016}]%
        {zhu2016generative}
\bibfield{author}{\bibinfo{person}{Jun-Yan Zhu}, \bibinfo{person}{Philipp
  Kr{\"a}henb{\"u}hl}, \bibinfo{person}{Eli Shechtman}, {and}
  \bibinfo{person}{Alexei~A Efros}.} \bibinfo{year}{2016}\natexlab{}.
\newblock \showarticletitle{Generative visual manipulation on the natural image
  manifold}. In \bibinfo{booktitle}{\emph{European Conference on Computer
  Vision}}. Springer, \bibinfo{pages}{597--613}.
\newblock


\end{thebibliography}

\clearpage
\appendix
\appendixpage
\section{Temporal coherent sequences}
As presented in Section \ref{subsec:applications} in the main paper, our method can generate temporally coherent sequences. To test the stability of the generated identity, we perturb facial attributes through a driving sequence of a facial performance. \figref{fig:videos} depicts how the identity is preserved almost to perfection in the presence of varying driving attributes, even though each image is generated independently. 
The sequences along each row are generated using the same driving sequence, and hence portray corresponding facial expression and pose. Note how the identity remains constant along the entire sequence, and how the expression, illumination and pose attributes match well along each row. The successful temporal coherence in the above experiment holds a promising avenue for future research in one-shot face reenactment.

\section{Interpolation in $\mathcal{Z}$}
\label{sec:inter_z}

\begin{figure*}[]
	\centering
	\includegraphics[width=0.9\linewidth]{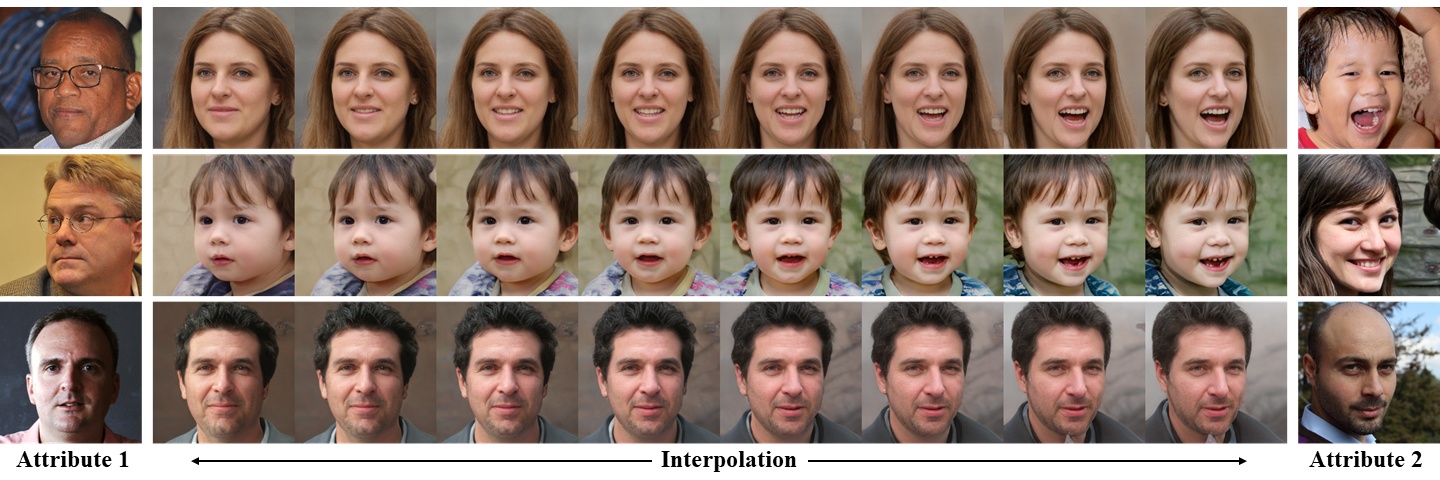}
	\caption{Disentangled interpolation of attributes while preserving identity. In each line, attributes representations are extracted from two images (both ends of the spectrum), and the identity representation is extracted from a third image (not shown). We interpolate the attributes representation, between the two extracted value. We then concatenate the interpolated attributes value with the fixed identity representation, forming $z$, a new whole latent face representation which is then used to generate an image (middle).}
	\label{fig:z_inter_attr}
	
	\vspace{0.5cm}
	
	\includegraphics[width=0.9\linewidth]{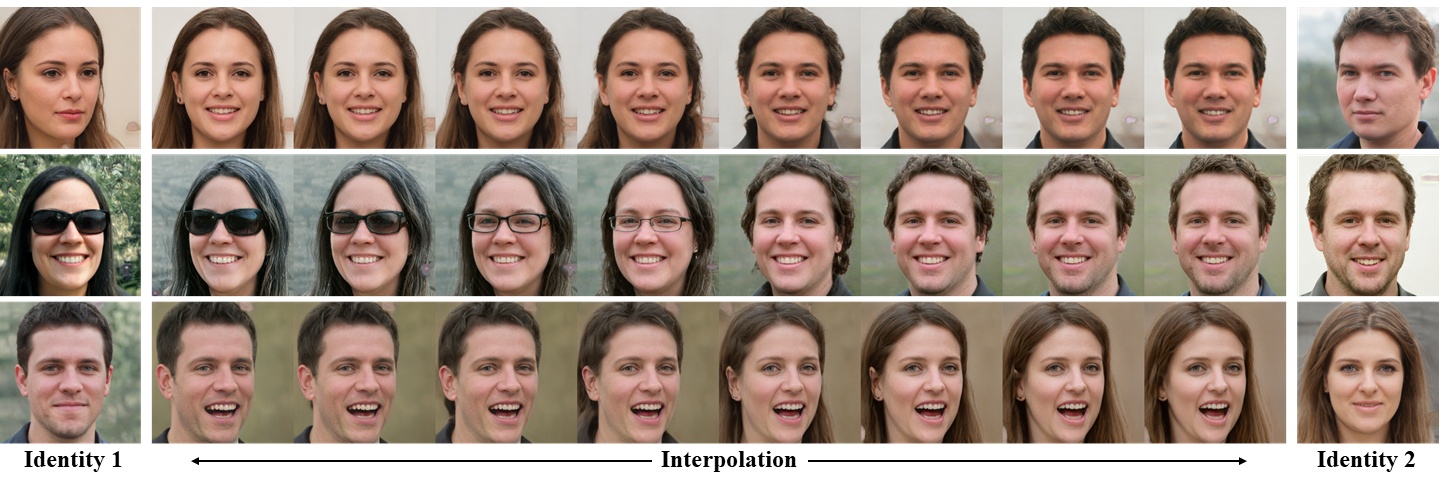}
	\caption{Disentangled interpolation of identity while preserving the other attributes. The setting is identical to the one in \figref{fig:z_inter_attr}, only here the attributes are extracted from the same image (not shown), and the identity is extracted from two images (both ends of the spectrum).}
	\label{fig:z_inter_id}
\end{figure*}

In \Cref{fig:z_inter_attr,fig:z_inter_id} we present the results of an experiment regarding the disentangled interpolation in $\mathcal{Z}$. This is different than the disentangled interpolation in $\mathcal{W}$, which we present in Section \ref{subsec:applications} in the paper. Equivalently to the latter figure, in each figure, the interpolated feature is extracted from the images on the two far ends, while the constant feature is extracted from a third image, which is not shown. These extracted features are our disentangled representations. We keep the representation of the constant feature fixed, while interpolating between the two representations of the other feature. The fixed and interpolated values are concatenated to form our latent code $z$. Our mapping networks converts this code to $w$ - a code in StyleGAN's latent space, which in turn is used to generate the image. The generated in-between images of each of these figures portray a pleasant and natural transition between the various explored properties, like they do in the parallel experiment in the paper. 
In \figref{fig:z_inter_attr}, we demonstrate that we accurately and consistently preserve the identity, while smoothly interpolating the expressions, poses and illumination, which are also well preserved from their respective inputs. In \figref{fig:z_inter_id}, we demonstrate that we accurately and consistently preserve the attributes, while smoothly interpolating the identity. 
Specifically, note the successful interpolation of illumination presented on the bottom row of \figref{fig:z_inter_attr} and the smooth disappearance of sunglasses, while maintaining high realism in the middle row of \figref{fig:z_inter_id}.

It is important to distinguish between interpolation in $\mathcal{W}$ and $\mathcal{Z}$. The well-behaved interpolation in $\mathcal{W}$ is inherited by StyleGAN, which is not the case for $\mathcal{Z}$, that is learned by our method. The fact that interpolating one factor representation in $\mathcal{Z}$, while keeping the other fixed, causes only a single factor to change in the generating image indicates that our representation is in fact disentangled.

\begin{frame}{ }
\begin{figure}[]
    \centering
    \animategraphics[autoplay,loop,width=0.9\linewidth]{15}{vid_frames/siga_videos-}{0}{180}
    \caption{Sample results for temporally coherent sequence generation experiment. For each identity, a single never-before-seen image was given, note the temporal coherency of identity along all sequences, and the coherency of the rest of the attributes along each row. 
    NOTE: this is an animated figure. Viewing it using Adobe Acrobat, or another animation supporting viewer, is strongly recommended.}
    \label{fig:videos}
\end{figure}
\end{frame}

\section{High Resolution Results}
\label{sec:HQ_results}
\begin{figure*}
	\centering
	\includegraphics[width=\linewidth]{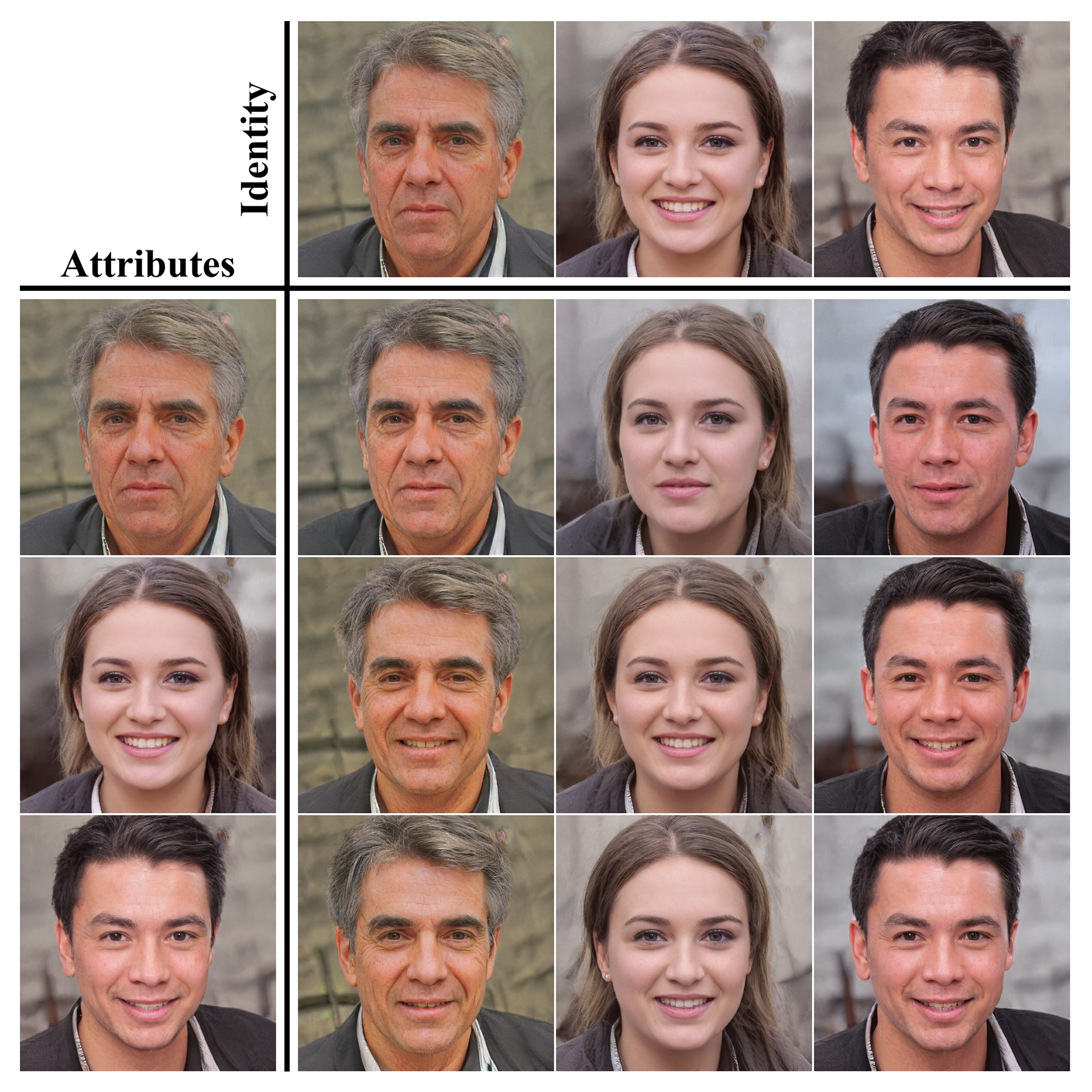}
	\caption{Feature combination results, in 1024x1024 resolution. For every image in the table, identity is taken from the top, and the rest of the attributes (including expressions, pose, lighting conditions, etc.) from the left. All images were generated using StyleGAN generator.}
	\label{fig:1024_table_1}
\end{figure*}

\begin{figure*}
	\centering
	\includegraphics[width=\linewidth]{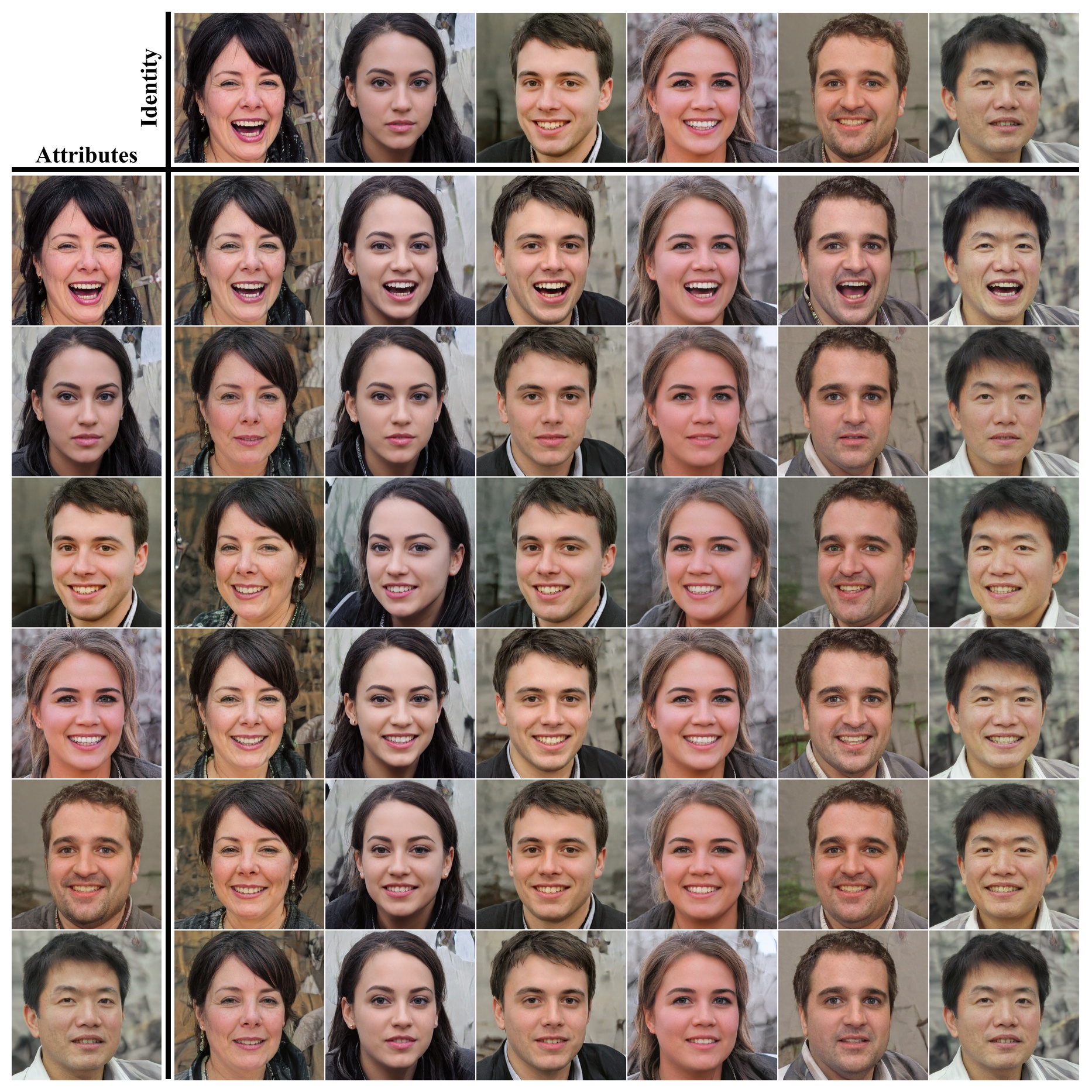}
	\caption{Feature combination results, in 1024x1024 resolution, as in \figref{fig:1024_table_1}.}
	\label{fig:1024_table_2}
\end{figure*}

\begin{figure*}
	\centering
	\includegraphics[width=\linewidth]{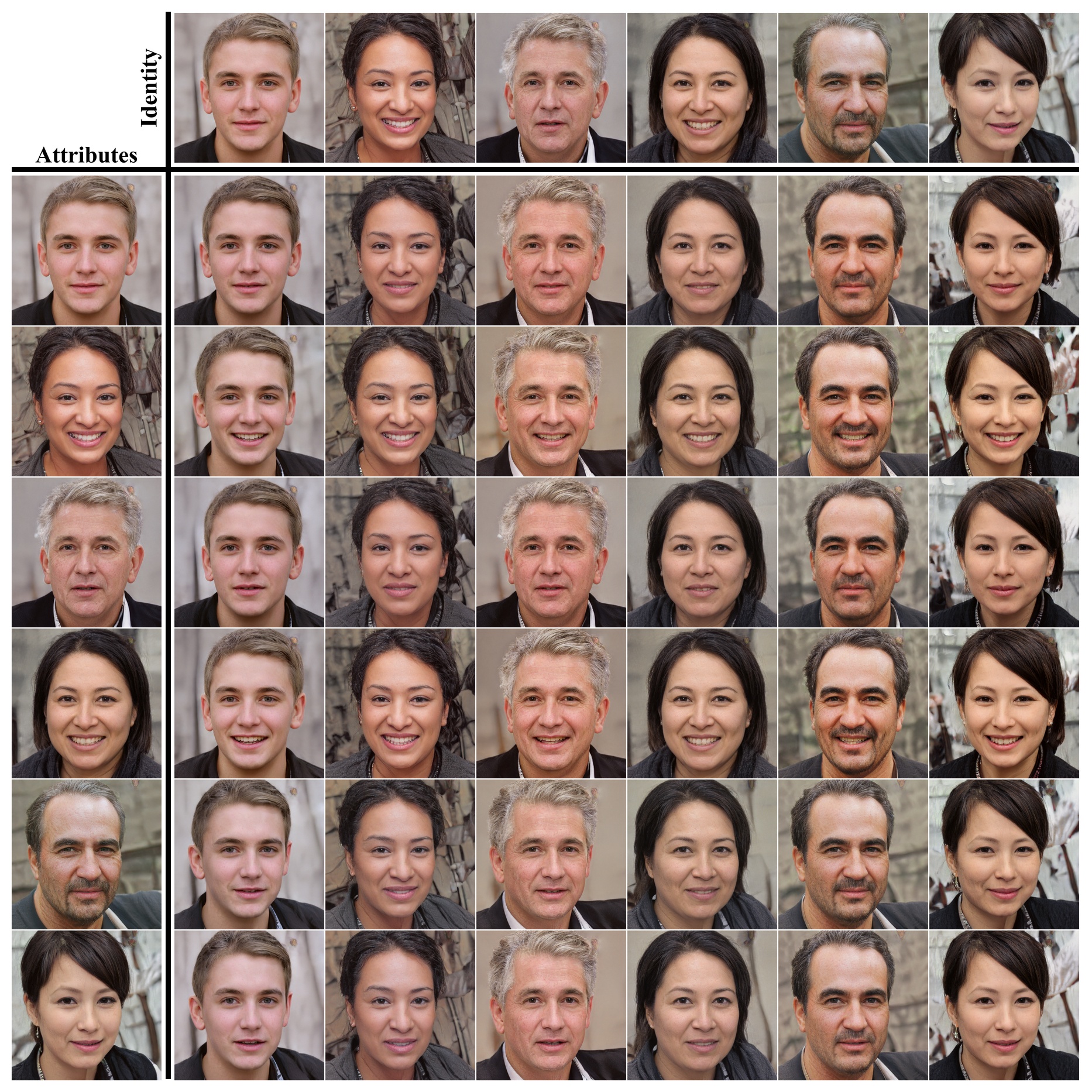}
	\caption{Feature combination results, in 1024x1024 resolution, as in \figref{fig:1024_table_1}.}
	\label{fig:1024_table_3}
\end{figure*}

\begin{figure*}
	\centering
	\includegraphics[width=\linewidth]{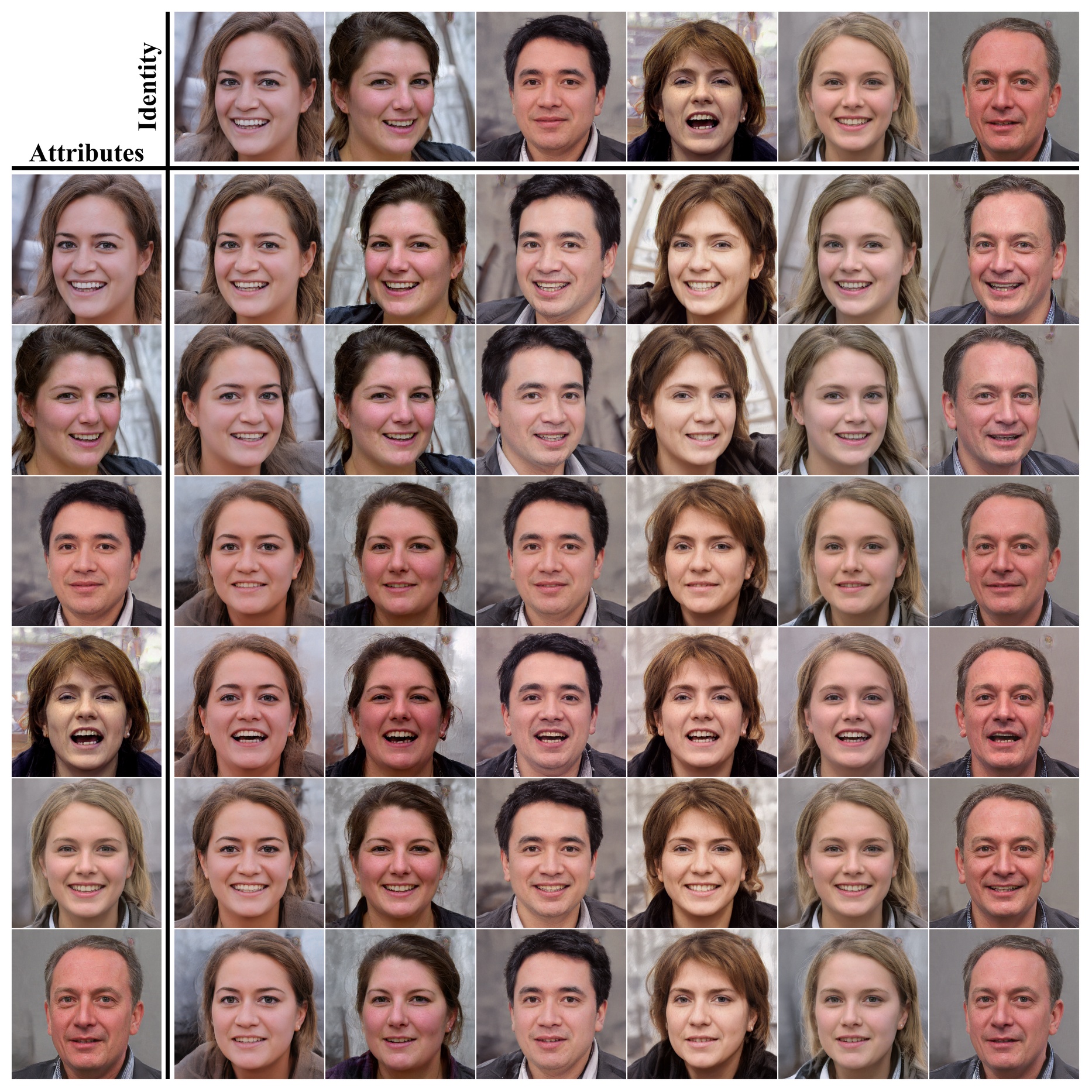}
	\caption{Feature combination results, in 1024x1024 resolution, as in \figref{fig:1024_table_1}.}
	\label{fig:1024_table_4}
\end{figure*}

All results in the main paper were obtained using a pre-trained StyleGAN with 256x256 resolution, to make results better comparable to previous methods. However, our method is able to fully harness the power of StyleGAN by generating 1024x1024 resolution.
We simply replace the pretrained 256x256 network with a pre-trained 1024x1024 one, and train in the same manner. In order to adjust to the higher memory requirements, we change the batch size to 2, and divide all learning rates by 10. The network is trained end-to-end on a single NVIDIA Titan XP GPU and requires roughly 3 days to converge. Note that this is incredibly efficient, compared to training StyleGAN itself on the same GPU, which requires roughly 60 days. Results are displayed in \Cref{fig:1024_table_1,fig:1024_table_2,fig:1024_table_3,fig:1024_table_4}.

\section{Results}

\begin{figure*}[t]
	\centering
	\includegraphics[width=\linewidth]{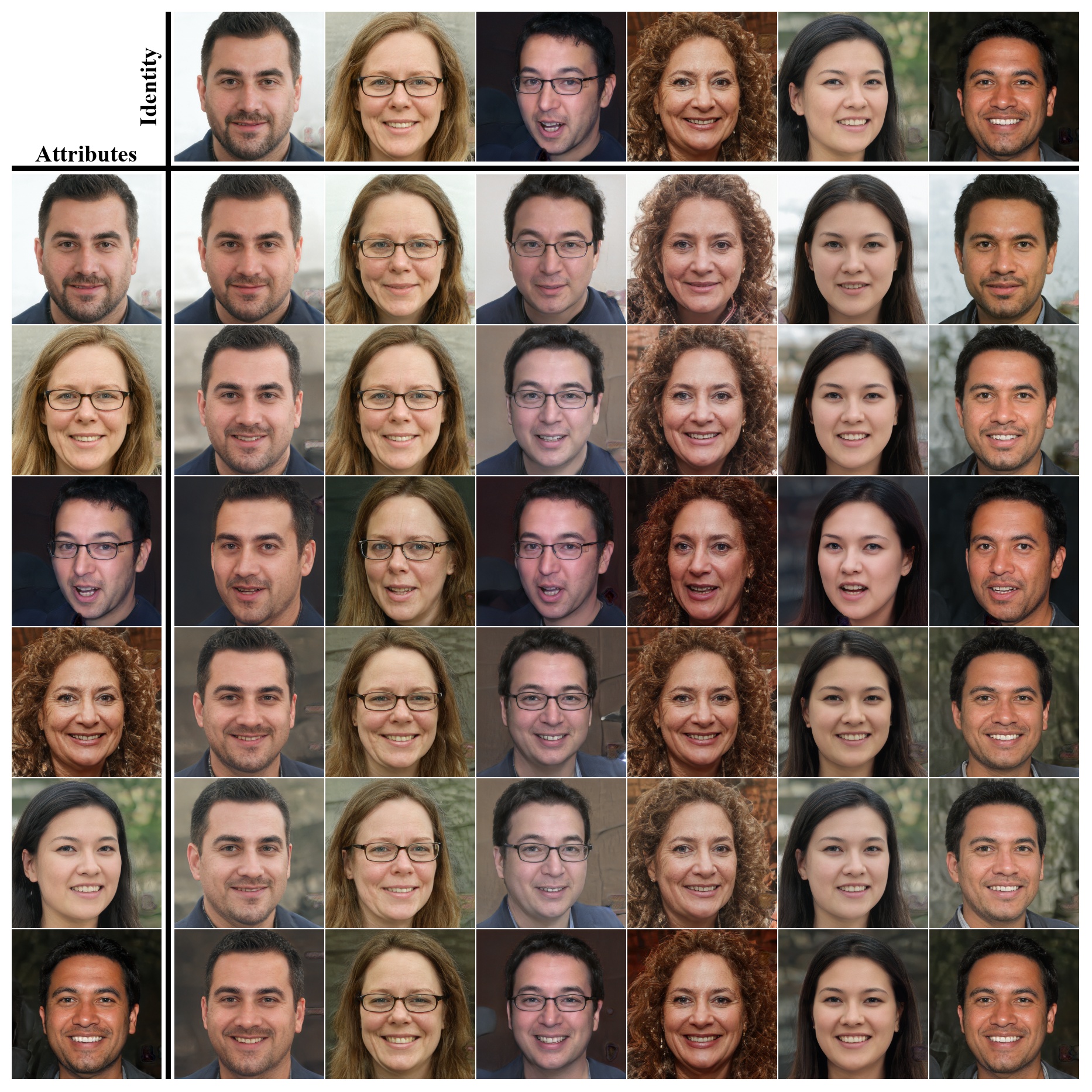}
	\caption{Further feature combination results, in 256x256 resolution. For every image in the table, identity is taken from the top, and the rest of the attributes (including expressions, pose, lighting conditions, etc.) from the left. All images were generated using StyleGAN.}
	\label{fig:256_table_2}
\end{figure*}

\begin{figure*}[t]
	\centering
	\includegraphics[width=\linewidth]{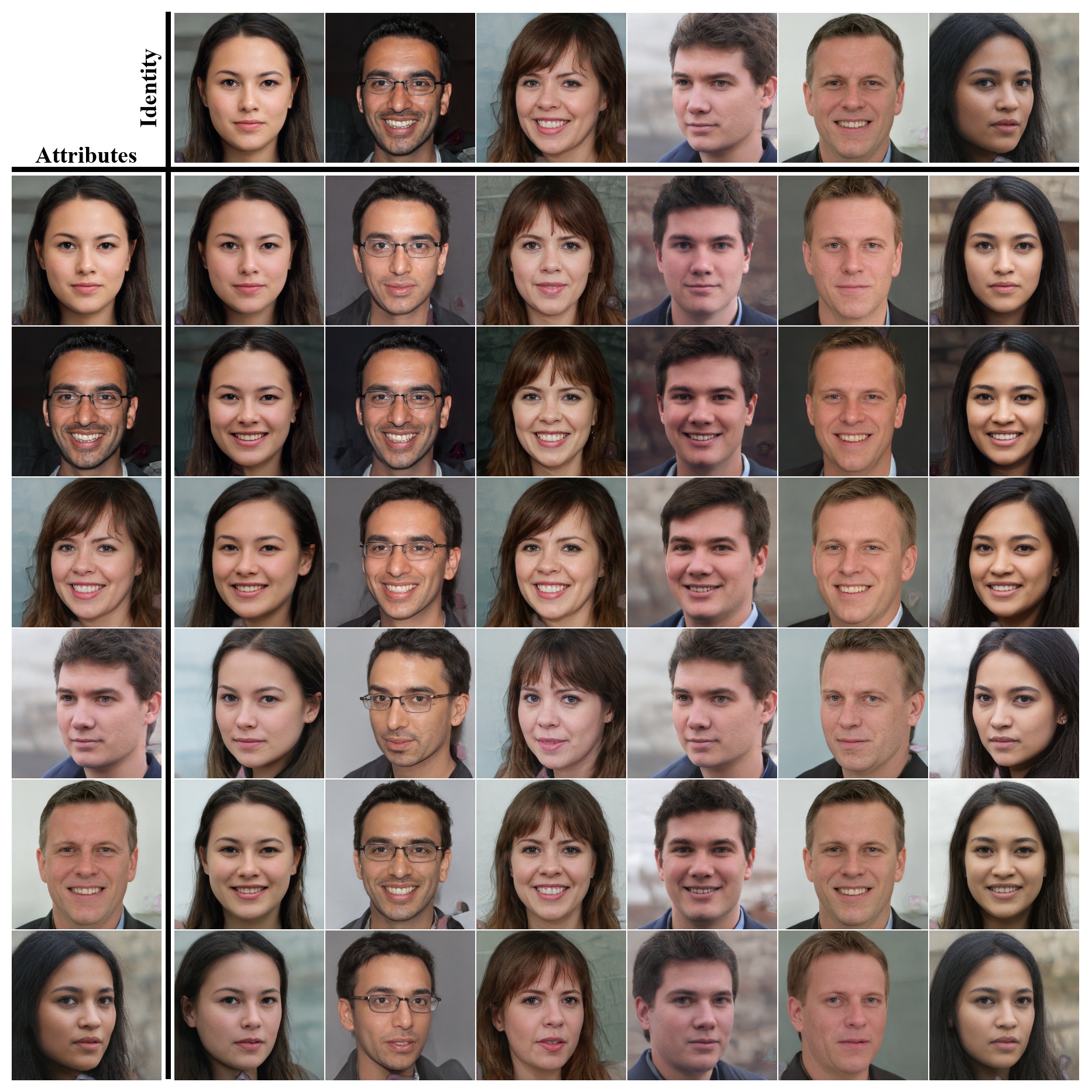}
	\caption{Further feature combination results, in 256x256 resolution, as in \figref{fig:256_table_2}}
	\label{fig:256_table_1}
\end{figure*}

We provide more 256x256 results, generated by our method in \Cref{fig:256_table_1,fig:256_table_2}.

\begin{figure*}
	\centering
	\includegraphics[width=\linewidth]{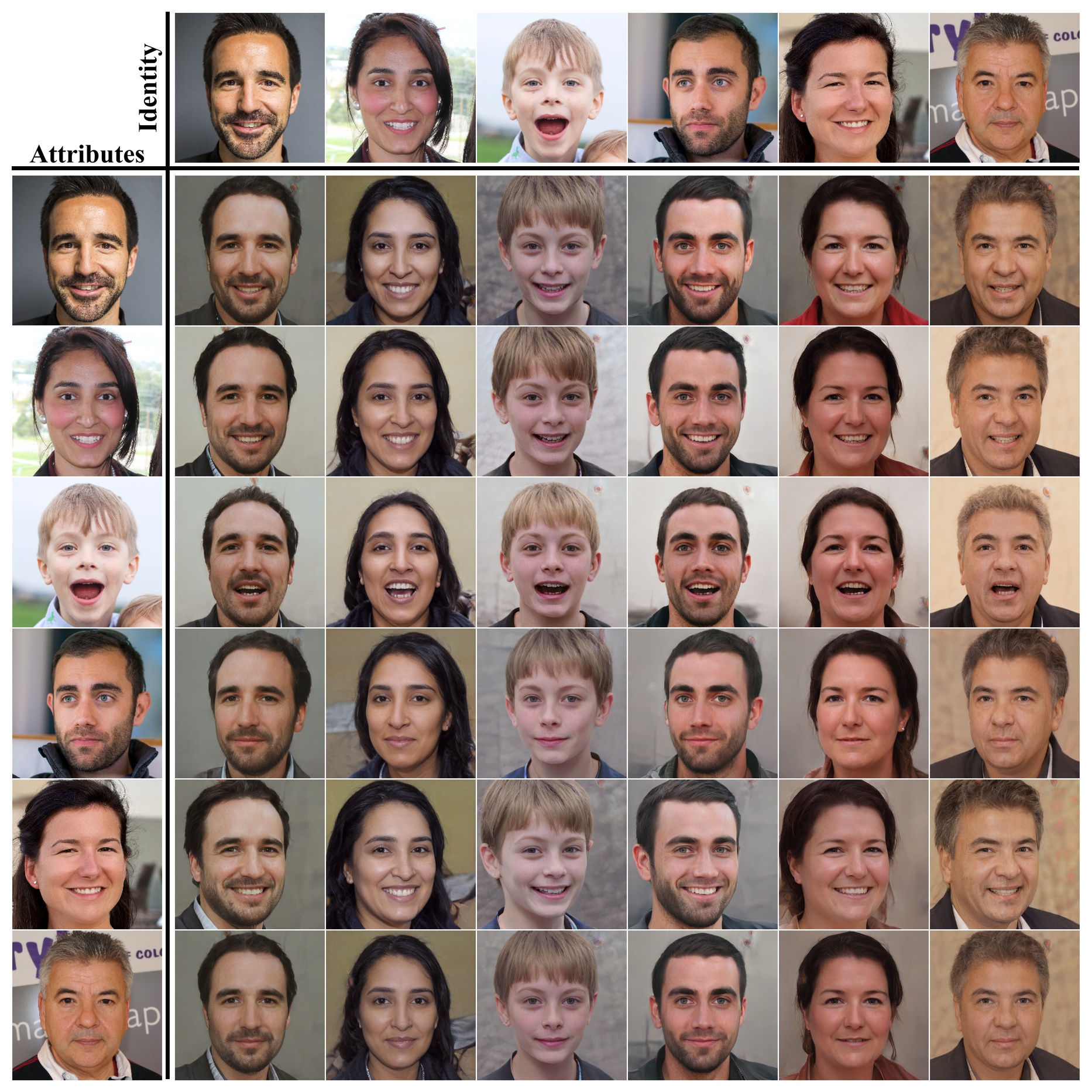}
	\caption{Feature combination results, in 1024x1024 resolution. The setting is identical to \ref{fig:1024_table_1}, only here input images are sampled from FFHQ}
	\label{fig:1024_ffhq_table_1}
\end{figure*}

\end{document}